%% file: main.tex
\documentclass{article}
\usepackage{arxiv}
\usepackage[utf8]{inputenc} 
\usepackage[T1]{fontenc}   
\usepackage{hyperref}    
\usepackage{url}        
\usepackage{booktabs}   
\usepackage{amsfonts}     
\usepackage{nicefrac}  
\usepackage{microtype}   
\usepackage{color}
\usepackage{xcolor}        
\usepackage{tcolorbox}
\usepackage{graphicx}
\usepackage{subfigure}
\usepackage{wrapfig}

\usepackage{amsmath}
\usepackage{amssymb}
\usepackage{mathtools}
\usepackage{amsthm}
\usepackage{multirow}
\usepackage[capitalize,noabbrev]{cleveref}
\theoremstyle{plain}
\newtheorem{theorem}{Theorem}[section]

\newtheorem{lemma}[theorem]{Lemma}
\newtheorem{corollary}[theorem]{Corollary}
\theoremstyle{definition}
\newtheorem{definition}[theorem]{Definition}

\theoremstyle{remark}

\usepackage{accents}
\usepackage[textsize=tiny]{todonotes}

\usepackage[utf8]{inputenc}
\usepackage[english]{babel}
\usepackage{mathrsfs}
\usepackage{times}
\usepackage{latexsym}
\usepackage{comment}

\usepackage[]{mathtools}   
\def\##1\#{\begin{align}#1\end{align}}
\def\$#1\${\begin{align*}#1\end{align*}}
\usepackage{enumitem}
\usepackage[]{amsthm} 
\usepackage[]{amssymb}
\usepackage{bbm}
\usepackage{xcolor}
\usepackage{colortbl}
\definecolor{best}{HTML}{BAFFCD}
\definecolor{issue}{HTML}{FFC8BA}
\definecolor{bad}{HTML}{FFC87C}
\newcommand{\good}[1]{\cellcolor{best}#1} 
\newcommand{\bad}[1]{\cellcolor{issue}#1}

\DeclareMathOperator*{\argmin}{argmin}
\input{src/math}
\newcommand{\BR}{\bm{1}}

\newcommand{\f}{f^*}
\newcommand{\D}{\mathcal D}
\newcommand{\x}{x}
\newcommand{\ny}{\tilde{y}}
\newcommand{\nY}{\widetilde{Y}}

\usepackage{xcolor}

\ifodd 1

\newcommand{\clar}[1]{\textbf{\color{green}(NEED CLARIFICATION: #1)}}
\newcommand{\response}[1]{\textbf{\color{magenta}(RESPONSE: #1)}}
\else

\newcommand{\com}[1]{}
\newcommand{\clar}[1]{}
\newcommand{\response}[1]{}
\fi
\newcommand{\RNum}[1]{\uppercase\expandafter{\romannumeral #1\relax}}

\title{To Aggregate or Not?
Learning with Separate Noisy Labels}

\author{Jiaheng Wei\thanks{Equal contributions. } \\
UC Santa Cruz 
\And
Zhaowei Zhu$^*$\\
UC Santa Cruz \\
\And
Tianyi Luo\\
UC Santa Cruz \\
\And
Ehsan Amid \\
Google Brain
\And
Abhishek Kumar\\
Google Brain \\
\And
Yang Liu \thanks{Correspondence to yangliu@ucsc.edu. (Paper under review)}\\
UC Santa Cruz\\
}
\begin{document}

\maketitle

\begin{abstract}
The rawly collected training data often comes with separate noisy labels collected from multiple imperfect annotators (e.g., via crowdsourcing). A typically way of using these separate labels is to first aggregate them into one and apply standard training methods. The literature has also studied extensively on effective aggregation approaches. This paper revisits this choice and aims to provide an answer to the question of whether one should aggregate separate noisy labels into single ones or use them separately as given. We theoretically analyze the performance of both approaches under the empirical risk minimization framework for a number of popular loss functions, including the ones designed specifically for the problem of learning with noisy labels. 
Our theorems conclude that label separation is preferred over label aggregation when the noise rates are high, or the number of labelers/annotations is insufficient. Extensive empirical results validate our conclusions.

\end{abstract}

\input{src/introduction}
\input{src/formulation}

\input{src/basic}

\input{src/comparison}

\input{src/experiment}

\section{Conclusions}

When learning with separate noisy labels, we explore the answer to the question ``whether one should aggregate separate noisy labels into single ones or use them separately as given''. In the empirical risk minimization framework, we theoretically show that label separation could be more beneficial than label aggregation when the noise rates are high or the number of labelers is insufficient. These insights hold for a number of popular loss function including several robust treatments. Empirical results on synthetic and real-world datasets validate our conclusion. 

\bibliography{library,myref,noise_learning}
\bibliographystyle{plain}

\newpage

\appendix

\newpage

\input{src/appendices}

\input{src/proofs}

\end{document}

%% file: src/math.tex
\usepackage{amsmath,amsfonts,bm,amsthm,dsfont}

\def\eqref#1{equation~\ref{#1}}

\def\1{\bm{1}}
\def\0{\bm{0}}

\DeclareMathAlphabet{\mathsfit}{\encodingdefault}{\sfdefault}{m}{sl}
\SetMathAlphabet{\mathsfit}{bold}{\encodingdefault}{\sfdefault}{bx}{n}

\newcommand{\PP}{\mathbb P}

\usepackage{tcolorbox}
\definecolor{grey}{rgb}{0.33, 0.33, 0.33}

\newcommand{\squishlist}{
\begin{list}{{{\small{$\bullet$}}}}
{\setlength{\itemsep}{1pt}      \setlength{\parsep}{0pt}
\setlength{\topsep}{-2pt}       \setlength{\partopsep}{0pt}
\setlength{\leftmargin}{1em} \setlength{\labelwidth}{1em}
\setlength{\labelsep}{0.5em} } }
\newcommand{\squishend}{  \end{list}  }

\makeatletter
\renewcommand*\env@matrix[1][*\c@MaxMatrixCols c]{%
  \hskip -\arraycolsep
  \let\@ifnextchar\new@ifnextchar
  \array{#1}}
\makeatother

\def\tr{\mathop{\text{tr}}\kern.2ex}

\long\def\comment#1{}

\def\tr{\mathop{\text{Tr}}}

\newcommand{\bel}{\begin{eqnarray}\label}
\newcommand{\eel}{\end{eqnarray}}
\newcommand{\bes}{\begin{eqnarray*}}
\newcommand{\ees}{\end{eqnarray*}}

\def\sep{{\circ}}
\def\agg{{\bullet}}
\def\uni{u}
\def\pb{\overline{\Delta}_R}
\def\Rk{\mathfrak{R}}
\def\hf{\hat{f}}
\def\nD{\widetilde{\mathcal{D}}}
\def\nDe{\widetilde{D}}
\def\P{\mathbb P}
\def\E{\mathbb E}
\def\Bias{\text{Bias}}
\def\hr{\hat{R}}
\def\slc{\leftarrow}

\def\spl{\looparrowright}



%% file: src/introduction.tex
\section{Introduction}

Training high-quality deep neural networks for classification tasks typically requires a large quantity of annotated data. The raw training data often comes with separate noisy labels collected from multiple imperfect annotators. For example, the popular data collection paradigm crowdsourcing \cite{estelles2012towards,howe2006rise,liu2015online} offers the platform to collect such annotations from unverified crowd; medical records are often accompanied with diagnosis from multiple doctors \cite{albarqouni2016aggnet,setio2017validation}; news articles can receive multiple checkings (of the article being fake or not) from different experts \cite{mitra2015credbank,pennycook2019fighting}. This leads to the situation considered in this paper: learning with multiple separate noisy labels. 

The most popular approach to learning from the multiple separate labels would be aggregating the given labels for each instance \cite{raykar2010learning,whitehill2009whose,rodrigues2014gaussian,rodrigues2017learning,luo2019machine}, through an expectation-maximization (EM) inference technique. Each instance will then be provided with one single label, and applied with the standard training procedure.

The primary goal of this paper is to revisit the choice of aggregating separate labels and hope to provide practitioners understandings for the following question:
\[\textbf{Should the learner aggregate separate noisy labels for one instance into a single label or not?}\]
Our main contributions can be summarized as follows:
\squishlist
\item We provide theoretical insights on how separation methods and aggregation ones result in different biases (Theorem~\ref{tm:bounds_ce},~\ref{tm:bounds},~\ref{tm:bounds_pl}) and variances (Theorem~\ref{thm:var_l},~\ref{thm:var_lc},~\ref{thm:var_pl}) of the output classifier from training. Our analysis considers both the standard loss functions in use, as well as popular robust losses that are designed for the problem of learning with noisy labels.
\item By comparing the analytical proxy of the worst-case performance bounds, our theoretical results reveal that separating multiple noisy labels is preferred over label aggregation when the noise rates are high, or the number of labelers/annotations is insufficient. The results are consistent for both basic loss function $\ell$ and robust designs, including loss correction and peer loss. 
\item We carry out extensive experiments using both synthetic and real-world datasets to validate our theoretical findings.
\squishend

\subsection{Related Works}
\paragraph{Label separation vs label aggregation}
Existing works mainly compare the separation with aggregation by empirical results. For example, it has been shown that label separation could be effective in improving model performance and may be potentially more preferable than aggregated labels through majority voting~\cite{ipeirotis2014repeated}. When training with the cross-entropy loss, Sheng et.al~\cite{sheng2017majority} observes that label separation reduces the bias and roughness, and outperforms majority-voting aggregated labels. However, it is unclear whether the results hold when robust treatments are employed. Similar problems have also been studied in corrupted label detection with a result leaning towards separation but not proved \cite{zhu2022detect}. Another line of approach concentrates on the end-to-end training scheme or ensemble methods which takes all the separate noisy labels as the input during the training process \cite{zhou2012ensemble,guan2018said,rodrigues2018deep,chen2020structured,wei2022deep}, and learning from separate noisy labels directly.

\paragraph{Learning with noisy labels}
Popular approaches in learning with noisy labels could be broadly divided into following categories, i.e., (1) Adjusting the loss on noisy labels by: using the knowledge of noise label transition matrix \cite{natarajan2013learning,patrini2017making,xia2019anchor,zhu2021clusterability,zhu2022beyond}; re-weighting the per-sample loss by down-weighting instances with potentially wrong labels \cite{liu2016classification,chang2017active,bar2021multiplicative,majidi2021exponentiated,kumar2021constrained}; or refurbishing the noisy labels \cite{reed2014training,lukasik2020does,wei2021understanding}. (2) Robust loss designs that do not require the knowledge of noise transition matrix \cite{wang2019symmetric,amid2019robust,wang2021policy,ma2020normalized,liu2020peer,zhu2021second,wei2020optimizing}. (3) Regularization techniques to prevent deep neural networks from memorizing noisy labels \cite{xia2020robust,liu2020early,liu2022robust,cheng2021demystifying,wei2021open}. (4) Dynamical sample selection procedure which behaves like a semi-supervised manner and begins with a clean sample selection procedure, then makes use of the wrongly-labeled samples  \cite{liu2021adaptive,cheng2020learning,luo2020research}. For example, several methods \cite{han2018co,yu2019does,wei2020combating} adopt a mentor/peer network to select small-loss samples as ``clean'' ones for the student/peer network.
See~\cite{han2020survey,song2022learning} for a more detailed survey of existing noise-robust techniques.
 

%% file: src/formulation.tex
\section{Formulation}
Consider an $M$-class classification task and let $X\in \mathcal{X}$ and $Y\in \mathcal{Y} :=\{1,2,...,M\}$ denote the input examples and their corresponding labels, respectively.  
We assume that $(X, Y)\sim \mathcal D$, where $\mathcal D$ is the joint data distribution. Samples $(x,y)$ are generated according to random variables $(X, Y)$.  
In the clean and ideal scenario, the learner has access to $N$ training data points $D:=\{(\x_n,y_n)\}_{n \in [N]}$. Instead of having access to ground truth labels $y_n$s, we only have access to a set of noisy labels $\{\ny_{n,i}^{\sep}\}_{i\in[K]}$ for $n\in[N]$. For ease of presentation, we adopt the decorator $\sep$ to denote separate labels, and $\agg$ for aggregated labels specified later.
Noisy labels $\ny_n^{\sep}$s are generated according to the random variable $\nY^{\sep}$. We consider the class-dependent label noise transition \cite{liu2016classification,natarajan2013learning} where $\nY^{\sep}$ is generated according to a transition matrix $T^{\sep}$ with its entries defined as follows:
\[
T_{k,l}^{\sep}:= \PP(\nY^{\sep}=l|Y=k).
\]
Most of the existing results on learning with noisy labels have considered the setting where each $\x_n$ is paired with only one noisy label $\ny_n^{\sep}$. In practice, we often operate in a setting where each data point $\x_n$ is associated with multiple separate labels drawn from the same noisy label generation process \cite{feldman2020does,liu2021understanding}. We consider this setting and assume that for each $\x_n$, there are $K$ independent noisy labels $\ny_{n,1}^{\sep},...,\ny_{n,K}^{\sep}$ obtained from $K$ annotators. 


%% file: src/basic.tex
We are interested in two popular ways to leverage  multiple separate noisy labels:
\squishlist
    \item Keep the separate labels as separate and apply standard learning with noisy labels techniques to each of them. 
    \item Aggregate noisy labels into one label, and then apply standard learning with noisy data techniques.
\squishend
We will look into each of the above two settings separately and then answer the question: 
\[\text{\emph{``Should the learner aggregate multiple separate noisy labels or not?''}}\]
\subsection{Label Separation}

Denote the column vector $\PP_{\nY^{\sep}}:= [\PP(\nY^{\sep}=1), \cdots, \PP(\nY^{\sep}={M})]^\top$ as the marginal distribution of $\nY^{\sep}$.
Accordingly, we can define $\PP_{Y}$ for $Y$. Clearly, we have the relation: $\PP_{\nY^{\sep}} = T^{\sep} \cdot \PP_{Y},  \PP_{Y} = (T^{\sep})^{-1} \cdot \PP_{\nY^{\sep}}.$
Denote by $\rho_{1}^{\sep} := \P(\nY^{\sep} = 0|Y=1), \rho_{0}^{\sep}:=\P(\nY^{\sep}=1|Y=0).$
The noise transition matrix $T$ has the following form when $M=2$:
\[T^{\sep}=\begin{bmatrix}
    1-\rho_0^{\sep} & \rho_0^{\sep}  \\
    \rho_1^{\sep} & 1-\rho_1^{\sep}  
    \end{bmatrix}.\]
For label separation, we define the per-sample loss function as:
\begin{align*}
    \ell(f(\x_n), \ny_{n,1}^{\sep},...,\ny_{n,K}^{\sep})=\frac{1}{K}\sum_{i\in [K]}\ell(f(\x_n),\ny_{n,i}^{\sep}).
\end{align*}
For simplicity, we shorthand $\ell(f(\x_n), \ny_{n}^{\sep}):=\ell(f(\x_n), \ny_{n,1}^{\sep},...,\ny_{n,K}^{\sep})$ for the loss of label separation method when there is no confusion. 
\subsection{Label Aggregation}
\begin{figure}[!htb]
 \vspace{-0.1in}
  \begin{center}
     \includegraphics[width=0.5\textwidth]{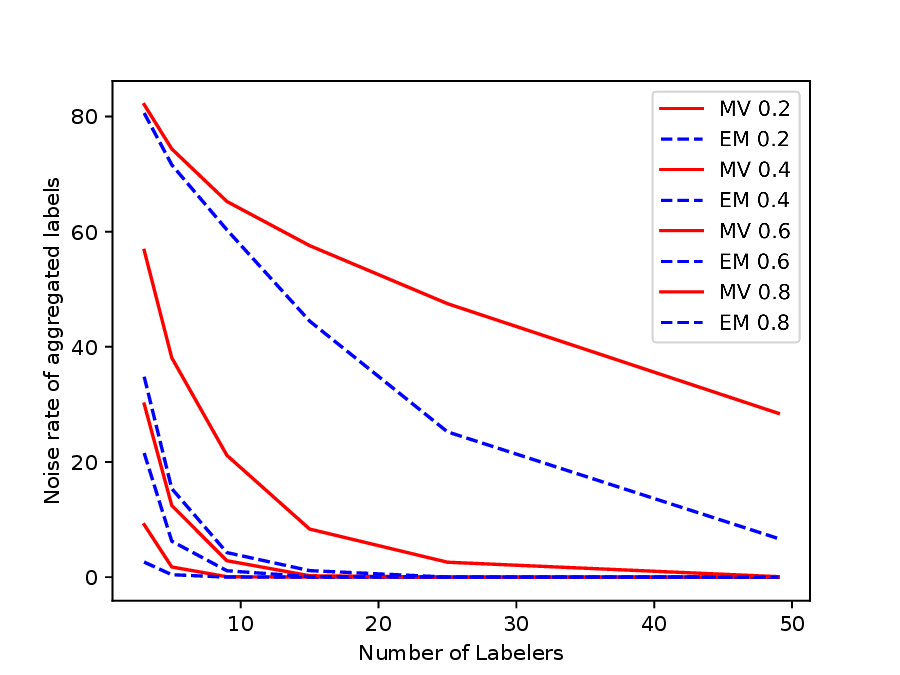}
  \end{center}
  \caption{Noise rates of the aggregated labels in synthetic noisy CIFAR-10. MV: majority vote. EM: expectation maximization. 0.2--0.8: Original noise rates before aggregation.
  }
    \label{fig:Aggregated_noise_rates}
\end{figure}

The other way to leverage multiple separate noisy labels is generating a single label via label aggregation methods using $K$ noisy ones:
\[
\ny^{\agg}_n := \textsf{Aggregation} (\ny_{n,1}^{\sep},\ny_{n,2}^{\sep},...,\ny_{n,K}^{\sep}),
\]
where the aggregated noisy labels $\ny_n^{\agg}$s are generated according to the random variable $\nY^{\agg}$. Denote the confusion matrix for this single \& aggregated noisy label as $T^{\agg}$.
 Popular aggregation methods include majority vote and EM inference, which are covered by our theoretical insights since our analyses in later sections would be built on the general label aggregation method. For a better understanding, we introduce the majority vote as an example.
\noindent \paragraph{Example of Majority Vote}
Given the majority voted label, we could compute the transition matrix between $\nY^{\agg}$ and the true label $Y$ using the knowledge of $T^{\sep}$.  The lemma below gives the closed form for $T^{\agg}$ in terms of $T^{\sep}$, when adopting majority vote.
\begin{lemma}\label{lm:agg}
Assume $K$ is odd and recall that in the binary classification task, $T_{i,j}^{\sep}=\PP(\nY^{\sep}=j|Y=i)$, the noise transition matrix of the (majority voting) aggregated noisy labels $T_{p,q}^{\agg}$ becomes:
\[
T_{p,q}^{\agg} = \sum_{i=0}^{\frac{K+1}{2}-1}{\binom{K}{i}(T^{\sep}_{p,q})^{K-i} (T^{\sep}_{p,1-q})^i},\quad p,q\in\{0, 1\}.
\]
\end{lemma}
When $K=3$, then ${T^{\agg}_{1,0}} = \PP(\nY^{\agg}=0|Y=1)=(T^{\sep}_{1,0})^3 + \binom{3}{1}(T^{\sep}_{1,0})^2(T^{\sep}_{1,1})$. Note it still holds that ${T_{p,q}^{\agg}+T_{p,1-q}^{\agg}=1}$.
For the aggregation method, as illustrated in Figure \ref{fig:Aggregated_noise_rates},  the x-axis indicates the number of labelers $K$, and the y-axis denotes the aggregated noise rate given that the overall noise rate is in $[0.2, 0.4, 0.6, 0.8]$. When the number of labelers is large (i.e., $K<10$) and the noise rate is small, both majority vote and EM label aggregation methods significantly reduce the noise rate. Although the expectation maximization method consumes much more time when generating the aggregated label, it frequently results in a lower aggregated noise rate than majority vote.

%% file: src/comparison.tex
\section{Bias and Variance Analyses w.r.t. $\ell$-loss}\label{sec:ell_theory}

In this section, we provide theoretical insights on how label separation and aggregation methods result in different biases and variances of the classifier prediction, when learning with the standard loss function $\ell$.

Suppose the clean training samples $\{(\x_n,y_n)\}_{n \in [N]}$ are given by variables $(X, Y)$ such that $(X, Y)\sim \D$.
Recall that instead of having access to a set of clean training samples $D=\{(\x_n,y_n)\}_{n \in [N]}$, 
the learner only observes $K$ noisy labels $\ny_{n,1}^{\sep},...,\ny_{n,K}^{\sep}$ for each $\x_n$, denoted by $\nDe^{\sep}:=\{(\x_n, \ny_{n,1}^{\sep},...,\ny_{n,K}^{\sep})\}_{n \in [N]}$. For separation methods, the noisy training samples are obtained through variables $(X, \nY_1^{\sep})$, ..., $(X, \nY_K^{\sep})$ where $(X, \nY_i^{\sep})\sim \nD^{\sep}$ for $i\in [K]$. For aggregation methods such as majority vote, we assume the data points and aggregated noisy labels $\nDe^{\agg}:=\{(\x_n, \ny_{n}^{\agg})\}_{n \in [N]}$ are drawn from $(X, \nY^{\agg})\sim \nD^{\agg}$ where $\nY^{\agg}$ is produced through the majority voting of $\nY_1^{\sep}, ..., \nY_K^{\sep}$.  {When we mention "noise rate", it usually refers to the average noise: $\mathbb P(\widetilde Y^{\text{u}} \ne Y)$.}
\vspace{-0.1in}
\paragraph{$\ell$-risk under the distribution}

Given the loss $\ell$, note that $ \ell(f(\x_n), \ny_{n}^{\sep})$ is denoted as $\ell(f(\x_n), \ny_{n,1}^{\sep},...,\ny_{n,K}^{\sep})=\frac{1}{K}\sum_{i\in [K]}\ell(f(\x_n),\ny_{n,i}^{\sep})$, we define the empirical $\ell$-risk for learning with separated/aggregated labels under noisy labels as: $\hr_{\ell, \nDe^{\uni}}(f)=\frac{1}{N}\sum_{i=1}^{N}\ell\left(f(\x_i), {\ny^{\uni}_i}\right)$, $\uni \in \{\sep, \agg\}$ unifies the treatment which is either separation $\sep$ or aggregation $\agg$.
By increasing the sample size $N$, we would expect $\hr_{\ell, \nDe^{\uni}}(f)$ to be close to the following $\ell$-risk under the noisy distribution $\nD^{\uni}$: $R_{\ell,\nD^{\uni}}(f)=\mathbb{E}_{(X,{\nY^{\uni}})\sim {\nD^{\uni}}}[\ell(f(X), {\nY^{\uni}})]$.
\vspace{-0.1in}
\subsection{Bias of a Given Classifier w.r.t. $\ell$-Loss}
We denote by $f^*\in \mathcal{F}$ the optimal classifier obtained through the clean data distribution $(X, Y)\sim \D$ within the hypothesis space $\mathcal{F}$. We formally define the bias of a given classifier $\hf$ as:
\begin{definition}[Classifier Prediction Bias of $\ell$-Loss]
Denote by {\small$R_{\ell, \D}(\hf):=\E_{\D}[\ell(\hf(X),Y)]$, $R_{\ell, \D}(f^*):=\E_{\D}[\ell(f^*(X),Y)]$}.
The bias of classifier $\hf$ writes as: ${\small\Bias(\hf)=R_{\ell, \D}(\hf)-R_{\ell, \D}(f^*).}$
\end{definition}

The $\Bias$ term quantifies the prediction bias (excess risk) of a given classifier $\hf$ on the clean data distribution $\mathcal{D}$ w.r.t. the optimal achievable classifier $f^*$, which can be decomposed as \cite{zhu2021rich}
\begin{align}\label{eq:decompose-ell}
   \Bias(\hf)
   =&\underbrace{R_{\ell, \D}(\hf)-R_{\ell, \nD^{\uni}}(\hf)}_{\text{Distribution shift}}+\underbrace{R_{\ell,\nD^{\uni}}(\hf)-  R_{\ell, \D}(f^*) }_{\text{Estimation error}}.
\end{align}
Now we bound the distribution shift and the estimation error in the following two lemmas.

\begin{lemma}[Distribution shift]
\label{lm:distribution_shift}
Denote by $p_i:=\P(Y=i)$, assume $\ell$ is upper bounded by $\bar{\ell}$ and lower bounded by $\underline{\ell}$.
The distribution shift in Eqn.~(\ref{eq:decompose-ell}) is upper bounded by
\begin{align}
R_{\ell, \D}(\hf)-R_{\ell, \nD^{\uni}}(\hf) \le \pb^{\uni,1} := (\rho_0^{\uni} p_0+\rho_1^{\uni} p_1)\cdot\left(\overline{\ell}-\underline{\ell}\right).
\label{eqm:shift_ce}
\end{align}
\end{lemma}

\begin{lemma}[Estimation error]
\label{lm:estimation_error}
Suppose the loss function $\ell(f(x), y)$ is $L$-Lipschitz for any feasible $y$.
$\forall f \in \mathcal F,$ with probability at least $1-\delta$, the estimation error is upper bounded by 
\[
R_{\ell,\nD^{\uni}}(\hf)-R_{\ell, \D}(f^*) \le \pb^{\uni,2} := 4L\cdot \Rk(\mathcal{F})+(\overline{\ell}- \underline{\ell})\cdot\sqrt{\frac{2\log(1/\delta)}{{\eta^{\uni}_K}N}}+\pb^{\uni,1},
\]
where $\uni \in \{\sep, \agg\}$ denotes either separation or aggregation methods, $\eta^{\sep}_K=\frac{K\cdot \log(\frac{1}{\delta})}{2\left(\log(\frac{K+1}{\delta})\right)^2}$ and $\eta^{\agg}_K\equiv 1$ indicate the richness factor, which characterizes the effect of the number of labelers, and $\Rk(\mathcal{F})$
is the Rademacher complexity of $\mathcal{F}$.
\end{lemma}

\begin{figure}[!htb]
  \begin{center}
     \includegraphics[width=0.45\textwidth]{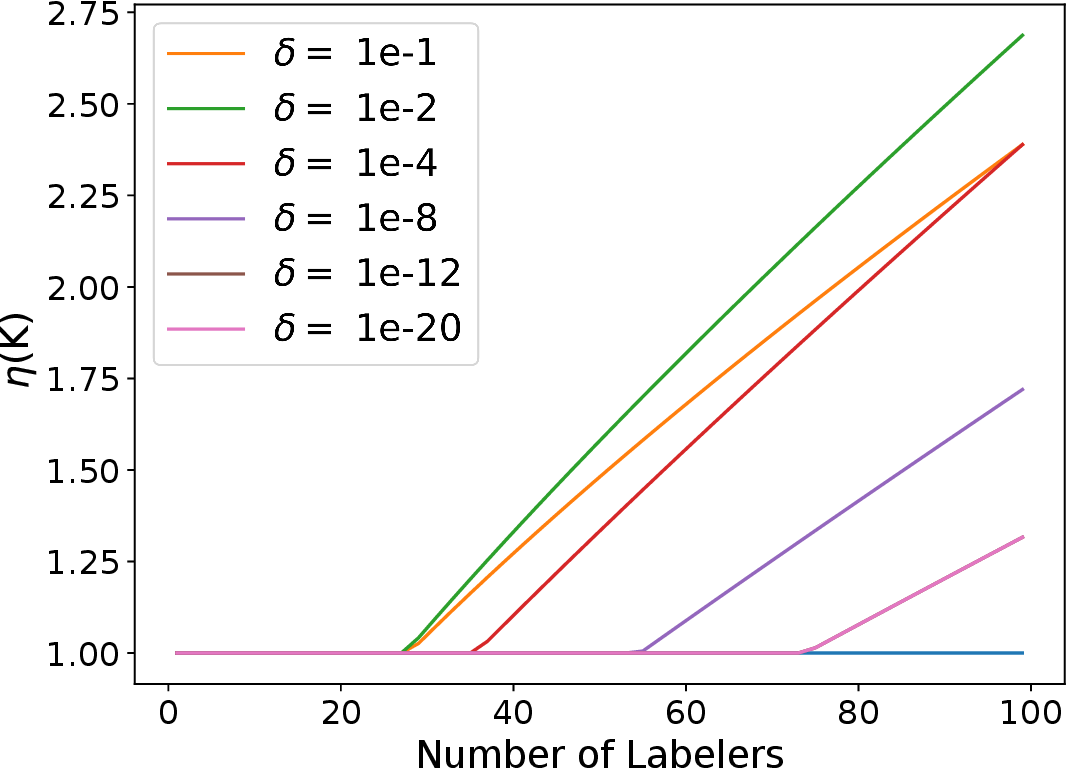}
  \end{center}
  \caption{The visualization of estimated $\eta_K^{\sep}$ given varied $\delta$.}
    \label{fig:etak}
\end{figure}

Noting that the number of unique instances $x_i$ are the same for both treatments, the duplicated copies of $x_i$ are supposed to introduce at least no less effective samples, i.e., the richness factor satisfies that $\eta_K^\uni\geq 1$. Thus, we update $
\eta_K^{\sep}:=\max\{\eta_K^{\sep}, 1\}$,
and Figure~\ref{fig:etak} visualizes the estimated $\eta_K^{\sep}$ given different number of labelers as well as $\delta$. It is clear that when the number of labelers is larger, or $\delta$ is smaller, $\eta_K^{\sep} > \eta_K^{\agg}$. Later we shall show how $\eta_K^{\uni}$ influences the bias and variance of the classifier prediction.

To give a more intuitive comparison of the performance of both mechanisms, we adopt the worst-case bias upper bound $\pb^{\uni}:=\pb^{\uni,1} + \pb^{\uni,2}$ from Lemma~\ref{lm:distribution_shift} and Lemma~\ref{lm:estimation_error} as a proxy and derive Theorem~\ref{tm:bounds_ce}.
\begin{theorem}\label{tm:bounds_ce}
Denote by $\alpha_K:=(\rho^{\sep}_0 p_0+\rho^{\sep}_1 p_1)-(\rho^{\agg}_0 p_0+\rho^{\agg}_1 p_1)$, $\gamma = \sqrt{{\log(1/\delta)}/{2N}}$.
  The separation bias proxy $\pb^{\sep}$ is smaller than the aggregation bias proxy $\pb^{\agg}$ if and only if
  \begin{align}\label{eqn:ce_bias}
  \alpha_K \cdot \frac{1}{1-(\eta^{\sep}_K)^{-\frac{1}{2}}}\leq \gamma.
  \end{align}
\end{theorem}
Note that $\alpha_K$ and $\eta^{\sep}_K$ are non-decreasing w.r.t. the increase of $K$, in Section \ref{sec:justify_cond}, we will explore how the LHS of Eqn. (\ref{eqn:ce_bias}) is influenced by $K$: a short answer is that the LHS of Eqn. (\ref{eqn:ce_bias}) is (generally) monotonically increasing w.r.t. $K$ when $K$ is small, indicating that Eqn. (\ref{eqn:ce_bias}) is easier to be achieved given fixed $\delta, N$ and a smaller $K$ than a larger one. 

\subsection{Variance of a Given Classifier w.r.t. $\ell$-Loss}
We now move on to explore the variance of a given classifier when learning with $\ell$-loss, prior to the discussion, we define the variance of a given classifier as:
\begin{definition}[Classifier Prediction Variance of $\ell$-Loss]
The variance of a given classifier $\hf$ when learned with separation ($\sep$) or aggregation ($\agg$) is defined as:
\[\text{Var}(\hf)=\E_{(X, \nY^{\uni})\sim \nD^{\uni}}\left[\ell(\hf(X),\nY^{\uni})- \E_{(X, \nY^{\uni})\sim \nD^{\uni}}[\ell(\hf(X),\nY^{\uni})]\right]^2.\]
\end{definition}
For $g(x)=x-x^2$, we derive the closed form of $\text{Var}$ and the corresponding upper bound as below.
\begin{theorem}\label{thm:var_l}
When {$\eta^{\uni}_K\geq \frac{2\log(1/\delta)}{N}$},  given $\ell$ is 0-1 loss, we have:  {\begin{align}
    \text{Var}(\hf^{\uni})=g(R_{\ell,\nD^{\uni}}(\hf^{\uni}))\leq \overbrace{g\left(\sqrt{\frac{2\log(1/\delta)}{\eta^{\uni}_K N}}\right)}^{\text{Variance proxy}}.\label{eqn:var_l}
\end{align}}
The variance proxy of $\text{Var}(\hf^{\sep})$ in Eqn. (\ref{eqn:var_l}) is  smaller than that of $\text{Var}(\hf^{\agg})$.
\end{theorem}
Theorem \ref{thm:var_l} provides another view to decide on the choices of separation and aggregation methods, i.e., the proxy of classifier prediction variance. To extend the theoretical conclusions w.r.t. $\ell$ loss to the multi-class setting, we only need to modify the upper bound of the distribution shift in Eqn. (\ref{eqm:shift_ce}), as specified in the following corollary.
\begin{corollary}[Multi-Class Extension ($\ell$-Loss)]\label{coro:multi-class_ce}
In the $M$-class classification case, the upper bound of the distribution shift in Eqn. (\ref{eqm:shift_ce}) becomes: 
\begin{align}
R_{\ell, \D}(\hf)-R_{\ell,\nD^{\uni}}(\hf) \le \pb^{\uni,1} :=\sum_{j\in[M]}p_j\cdot(1-T^{\uni}_{j,j})\cdot\left(\overline{\ell}-\underline{\ell}\right).\end{align}
\end{corollary}

\section{Bias and Variance Analyses with Robust Treatments}

Intuitively, the learning of noisy labels problem could benefit from more robust loss functions build upon the generic $\ell$ loss, i.e., backward correction (surrogate loss) \cite{natarajan2013learning,patrini2017making}, and peer loss functions \cite{liu2020peer}. We move on to explore the best way to learn
with multiple copies of noisy labels, when combined with existing robust approaches.

\subsection{Backward Loss Correction}

When combined with the backward loss correction approach ($\ell\to \ell_{\slc}$), the empirical $\ell$ risks become: $\hr_{\ell_{\slc}, \nDe^{\uni}}(f)=\frac{1}{N}\sum_{i=1}^{N}\ell_{\slc}(f(\x_i), \tilde{y}^{\uni}_i),$
where the corrected loss in the binary case is defined as
$$\ell_{\slc}(f(\x), \ny^{\uni})=\frac{(1-\rho_{1-\ny^{\uni}})\cdot\ell(f(\x), \ny^{\uni})-\rho_{\ny^{\uni}}\cdot \ell(f(\x),1-\ny^{\uni})}{1-\rho^{\uni}_0-\rho^{\uni}_1}.$$
\paragraph{Bias of given classifier w.r.t. $\ell_{\slc}$}
Suppose the loss function $\ell(f(\x), y)$ is $L$-Lipschitz for any feasible $y$. 
Define {\small $L_{\slc}^{\uni}:= L_{\slc0}^{\uni} \cdot L$, where $L_{\slc0}^{\uni}:=\frac{(1+|\rho^{\uni}_0-\rho^{\uni}_1|)}{1-\rho^{\uni}_0-\rho^{\uni}_1}$}. Denote by $R_{\ell, \D}(\hf)$ the $\ell$-risk of the classifier $\hf$ under the clean data distribution $\D$, with $\hf=\hf_{\slc}^{\uni}=\argmin_{f\in \mathcal{F}} \hr_{\ell_{\slc}, \nDe^{\uni}}(f)$. Lemma~\ref{lm:bound} gives the upper bound of classifier prediction bias when learning with $\ell_{\slc}$ via separation or aggregation methods.
\begin{lemma}
\label{lm:bound}
With probability at least $1-\delta$, we have: 
$$R_{\ell, \D}(\hf_{\slc}^{\uni})-R_{\ell, \D}(f^*) 
\leq  {\pb^{\uni}}_{\slc}:= 4L_{\slc}^{\uni} \cdot\Rk(\mathcal{F})+L_{\slc 0}^{\uni} \cdot (\overline{\ell}- \underline{\ell})\cdot\sqrt{\frac{2\log(1/\delta)}{\eta^{\uni}_K N}}.$$
\end{lemma}
Lemma \ref{lm:bound} offers the upper bound of the performance gap for the given classifier $f$ w.r.t the clean distribution $\mathcal{D}$, comparing to the minimum achievable risk.  We consider the bound ${\pb^{\uni}}_{\slc}$ as a proxy of the bias, and we are interested in the case where training the classifier separately yields a smaller bias proxy compared to that of the aggregation method, formally ${\pb^{\sep}}_{\slc}<{\pb^{\agg}}_{\slc}$. For any finite hypothesis class $\mathcal{F}\subset \{f: X\to \{0, 1\}\}$, and the sample set $S=\{\x_1, ..., \x_N\}$, denote by $d$ the VC-dimension of $\mathcal{F}$, we give conditions when training separately yields a smaller bias proxy.
\begin{theorem}\label{tm:bounds}
Denote by $\alpha_K:= 1-{ L_{\slc}^{\agg}}/{L_{\slc}^{\sep}} $, $\gamma = 1/\left(1+ {\frac{4L}{\overline{\ell}-\underline{\ell}} \sqrt{\frac{d\log(N)}{\log(1/\delta)}}}\right)$, where $d$ is the VC-dimension of $\mathcal F$.
For backward loss correction, the separation bias proxy ${\pb^{\sep}}_{\slc}$ is smaller than the aggregation bias proxy ${\pb^{\agg}}_{\slc}$ if and only if
\begin{align}
  \alpha_K \cdot \frac{1}{1-(\eta^{\sep}_K)^{-\frac{1}{2}}}\leq \gamma.\label{eqn:lc_bias}
\end{align} 
\end{theorem}
We defer our empirical analysis of the monotonicity of the LHS in Eqn. (\ref{eqn:lc_bias}) to Section \ref{sec:justify_cond} as well, which shares similar monotonicity behavior to learning w.r.t. $\ell$.

\paragraph{Variance of given classifiers with Backward Loss Correction}
Similar to the previous subsection, we now move on to check how separation and aggregation methods result in different variance when training with loss correction. 
\begin{theorem}\label{thm:var_lc}
When $L_{\slc0}^{\uni}(\eta_K^{\uni})^{-\frac{1}{2}}<\sqrt{\frac{N}{2(\overline{\ell}- \underline{\ell})^2\log(1/\delta)}}$, $\text{Var}(\hf_{\slc}^{\uni})$ (w.r.t. the 0-1 loss) satisfies:
 {\begin{align}
    \text{Var}(\hf^{\uni}_{\slc})=g(R_{\ell,\nD^{\uni}}(\hf^{\uni}_{\slc}))\leq \overbrace{g\left({L_{\slc0}^{\uni}\cdot (\overline{\ell}- \underline{\ell})}\cdot\sqrt{\frac{2\log(1/\delta)}{\eta_K^{\uni}N}}\right)}^{\text{Variance Proxy}}.\label{eqn:var_lc}
\end{align}}
The variance proxy of $\text{Var}(\hf^{\sep}_{\slc})$ in Eqn. (\ref{eqn:var_lc}) is  smaller than that of $\text{Var}(\hf^{\agg}_{\slc})$ if {\small$\sqrt{{\eta_K^{\sep}}}>\frac{L_{\slc}^{\sep}}{L_{\slc}^{\agg}}$}.
\end{theorem}
Moving a bit further, when the noise transition matrix is symmetric for both methods, the requirement {\small$\sqrt{\eta_K^{\uni}}>\frac{L_{\slc}^{\sep}}{L_{\slc}^{\agg}}$} could be further simplified as: $\sqrt{\eta_K^{\uni}}>\frac{L_{\slc}^{\sep}}{L_{\slc}^{\agg}}=\frac{1-\rho^{\agg}_0-\rho^{\agg}_1}{1-\rho^{\sep}_0-\rho^{\sep}_1}$. For a fixed $K$, a more efficient aggregation method decreases $\rho_i^{\agg}$, which makes it harder to satisfy this condition.

Recall $ L_{\slc}^{\uni} := L_{\slc0}^\uni \cdot L$, the theoretical insights of $\ell_{\slc}$ between binary case and the multi-class setting could be bridged by replacing $L_0^\uni$ with the multi-class constant specified in the following corollary.
\begin{corollary}[Multi-Class Extension ($\ell_{\slc}$-Loss)]\label{coro:multi-class_lc}
Given a diagonal-dominant transition matrix $T^{\uni}$, we have
\[
L_{\slc0}^{\uni} = \frac{2\sqrt{M}}{\lambda_{\min}(T^\uni)},
\]
where $\lambda_{\min}(T^\uni)$ denotes the minimal eigenvalue of the matrix $T^\uni$. Particularly, if $T^\uni_{ii} < 0.5, \forall i\in[M]$, we further have
\[
L_{\slc0}^{\uni} = \min \left\{\frac{1}{1- 2  e^\uni},\frac{2\sqrt{M}}{\lambda_{\min}(T^\uni)}\right\}, \quad \text{where}~~ e^\uni:=\max_{i\in[M]} (1-T_{ii}^{\uni}).
\]
\end{corollary}

\subsection{Peer Loss Functions}
Peer Loss function \cite{liu2020peer} is a family of loss functions that are shown to be robust to label noise, without requiring the knowledge of noise rates. Formally, $\ell_{\spl}(f(\x_i),\ny_i):=\ell(f(\x_i),\ny_i)-\ell(f(\x_{i}^1),\ny_i^2)$, where the second term checks on mismatched data samples with $(\x_i,\ny_i)$, $(\x_i^1,\ny_i^1)$, $(\x_i^2,\ny_i^2)$, which are randomly drawn from the same data distribution. When combined with the peer loss approach, i.e., $\ell\to \ell_{\spl}$, the two risks become: $ \hr_{\ell_{\spl}, \nDe^{\uni}}(f)=\frac{1}{N}\sum_{i=1}^{N}\ell_{\spl}(f(\x_i), \tilde{y}^{\uni}_i), \uni \in \{\sep, \agg\}.$ 
\paragraph{Bias of given classifier w.r.t. $\ell_{\spl}$}
Suppose the loss function $\ell(f(\x), y)$ is $L$-Lipschitz for any feasible $y$. 
Let {\small $L_{\spl0}^{\uni}:=1/(1-\rho^{\uni}_0-\rho^{\uni}_1), L_{\spl}^{\uni}:=L_{\spl0}^{\uni}\cdot L$} and 
$\hf_{\spl}^{\uni} = \argmin_{f\in \mathcal{F}} \hr_{\ell_{\spl}, \nDe^{\uni}}(f)$.
\begin{lemma}
\label{lm:bound_pl}
With probability at least $1-\delta$, we have:
{{
\begin{align*}
&R_{\ell, \D}(\hf_{\spl}^{\uni})- R_{\ell, \D}(f^*) \leq {\pb^{\uni}}_{\spl}:=8L_{\spl}^{\uni}\cdot \Rk(\mathcal{F})+L_{\spl0}^{\uni}\cdot \sqrt{\frac{2\log(4/\delta)}{ \eta^{\uni}_K N}}\cdot\left(1+2(\bar{\ell}-\underline{\ell})\right).
\end{align*}
}}
\end{lemma}
To evaluate the performance of a given classifier yielded by the optimization w.r.t. $\ell_{\spl}$, Lemma \ref{lm:bound_pl} provides the bias proxy ${\pb^{\uni}}_{\spl}$ for both treatments. Similarly, we adopt such a proxy to analyze which treatment is more preferable.

\begin{theorem}\label{tm:bounds_pl}
Denote by $\alpha_K:= 1-{ L_{\spl}^{\agg}}/{L_{\spl}^{\sep}} $, $\gamma =  \frac{1+2(\bar{\ell}-\underline{\ell})}{2L}\sqrt{\frac{\log(4/\delta)}{4d\log(N)} } $, where $d$ denotes the VC-dimension of $\mathcal F$.
For peer loss, the separation bias proxy ${\pb^{\sep}}_{\spl}$ is smaller than the aggregation bias proxy ${\pb^{\agg}}_{\spl}$ if and only if
\begin{align}
  \alpha_K \cdot \frac{1}{{ L_{\spl}^{\agg}}/{L_{\spl}^{\sep}}-(\eta^{\sep}_K)^{-\frac{1}{2}}}\leq \gamma. \label{eqn:pl_bias}
\end{align}
\end{theorem}
Note that the condition in Eqn. (\ref{eqn:pl_bias}) shares a similar pattern to that which appeared in the basic loss $\ell$ and $\ell_{\slc}$. We will empirically illustrate the monotonicity of its LHS in Section \ref{sec:justify_cond}.

\paragraph{Variance of given classifiers with Peer Loss}
We now move on to check how separation and aggregation methods result in different variances when training with peer loss. Similarly, we can obtain:

\begin{theorem}\label{thm:var_pl}
When {\small $\sqrt{\eta_K^{\uni}}\geq  \sqrt{\frac{2\log(4/\delta)}{N}}\cdot \left(1+2(\bar{\ell}-\underline{\ell})\right)$},  $\text{Var}(\hf_{\spl}^{\uni})$ (w.r.t. the 0-1 loss) satisfies:
 {\begin{align}
    \text{Var}(\hf^{\uni}_{\spl})=g(R_{\ell,\nD^{\uni}}(\hf^{\uni}_{\spl}))\leq \overbrace{g\left(L_{\spl0}^{\uni}\cdot\sqrt{\frac{\log(4/\delta)}{2 \eta^{\uni}_K N}}\cdot\left(1+2(\bar{\ell}-\underline{\ell})\right)\right)}^{\text{Variance proxy}}.\label{eqn:var_pl}
\end{align}}
The variance proxy of $\text{Var}(\hf^{\sep}_{\spl})$ in Eqn. (\ref{eqn:var_pl}) is  smaller than that of $\text{Var}(\hf^{\agg}_{\spl})$ if $\sqrt{\eta_K^{\sep}}\geq \frac{L_{\spl}^{\sep}}{L_{\spl}^{\agg}}$.
\end{theorem}
Theoretical insights of $\ell_{\spl}$ also have the multi-class extensions, we only need to generate $L_{\spl0}^\uni$ to the multi-class setting along with additional conditions specified as below:
\begin{corollary}[Multi-Class Extension ($\ell_{\spl}$-Loss)]\label{coro:multi-class_pl}
Assume $\ell_{\spl}$ is classification-calibrated in the multi-class setting, and the clean label $Y$ has equal prior $P(Y=j)=\frac{1}{M}, \forall j\in [M]$. For the uniform noise transition matrix  \cite{wei2020optimizing} such that $T^{\uni}_{j,i}=\rho_i^{\uni}, \forall j\in[M]$, we have: $L_{\spl0}^{\uni} =1/(1-\sum_{i\in[M]}\rho_i^{\uni}).$
\end{corollary}

\subsection{Analysis of the Theoretical Conditions}\label{sec:justify_cond}
\begin{figure}[!htb]
    \centering
    \includegraphics[width=0.32\textwidth]{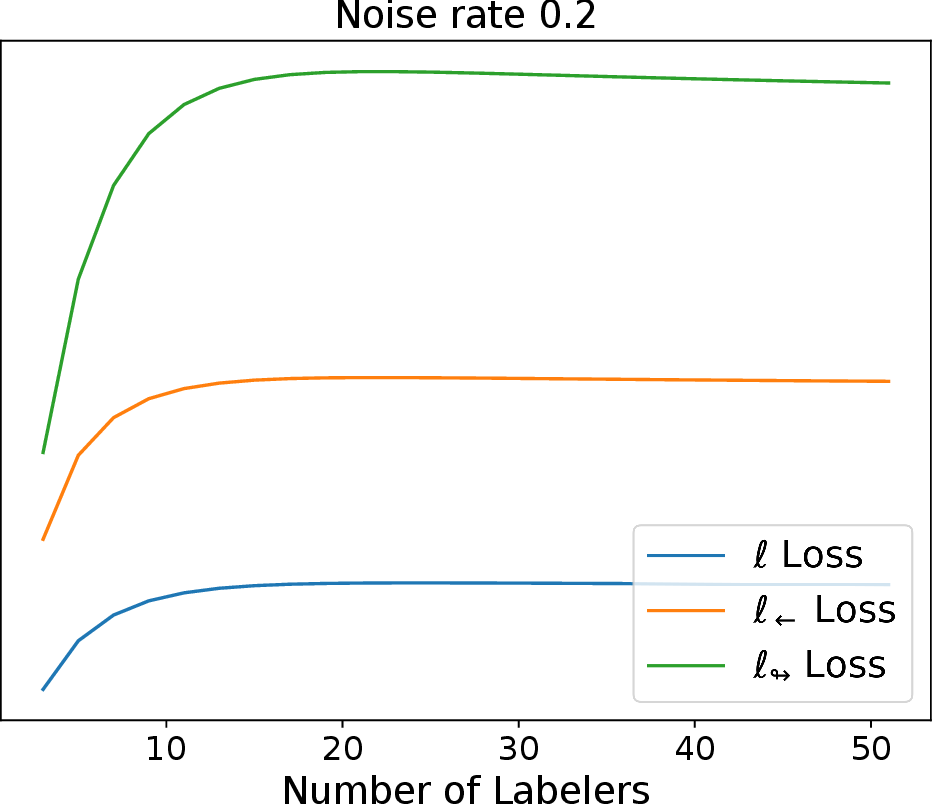}
    \hspace{0.5cm}
    \includegraphics[width=0.32\textwidth]{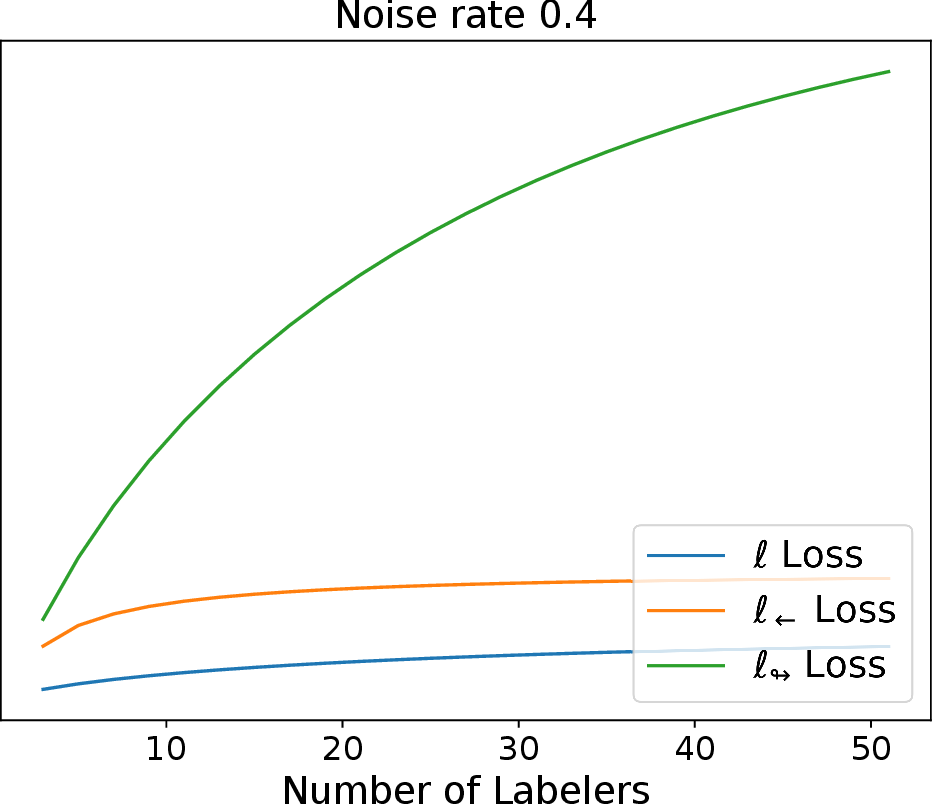}
     \caption{The monotonicity of the LHS in Eqn. (\ref{eqn:ce_bias}, \ref{eqn:lc_bias}, \ref{eqn:pl_bias}) w.r.t. the increase of $K$.}
    \label{fig:monotonicity}
    \end{figure}

 Recall that the established conditions in Theorems~\ref{tm:bounds_ce},~\ref{tm:bounds},~\ref{tm:bounds_pl} are implicitly relevant to the number of labelers $K$, and the RHS of Eqns.~(\ref{eqn:ce_bias}, \ref{eqn:lc_bias}, \ref{eqn:pl_bias}) are  constants. We proceed to analyze the monotonicity of the corresponding LHS (in the form of $\alpha_K\cdot \frac{1}{\beta_K-(\eta^{\sep}_K)^{-\frac{1}{2}}}$) w.r.t. the increase of $K$, where $\beta_K=1$ for $\ell$ and $\ell_{\slc}$, $\beta_K=L_{\spl}^{\agg}/L_{\spl}^{\sep}$ for $\ell_{\spl}$. Thus, we have: $O(\text{LHS})=O(\alpha_K\cdot (\beta_K-O(\frac{\log(K)}{\sqrt{K}}))^{-1})$. 
We visualize this order under different symmetric $T^{\sep}$ in Figure \ref{fig:monotonicity}. It can be observed that, when $K$ is small (e.g., $K\leq 5$), the LHS parts of these conditions increase with $K$, while they may decrease with $K$ if $K$ is sufficiently large.
Recall that separation is better if LHS is less than the constant value $\gamma$. Therefore, Figure~\ref{fig:monotonicity} shows the trends that aggregation is generally better than separation when $K$ is sufficiently large.

\paragraph{Tightness of the bias proxies}
In Theorems~\ref{tm:bounds_ce},~\ref{tm:bounds},~\ref{tm:bounds_pl}, we view the error bounds $\pb^{\uni}, {\pb^{\uni}}_{\slc}, {\pb^{\uni}}_{\spl}$ as proxies of the worst-case performance of the trained classifier. For the standard loss function $\ell$, it has been proven that \cite{mendelson2008lower,lecue2010sharper}  under mild conditions of $\ell$ and $\mathcal{F}$, the lower bound of the performance gap between a trained classifier ($\hf$) and the optimal achievable one (i.e., $f^*$) $R_{\ell, \D}(\hf)- R_{\ell, \D}(f^*)$ is of the order $O(\sqrt{{1}/{N}})$, which is of the same order as that in Theorem \ref{tm:bounds_ce}. 
Noting the behavior concluded from the worst-case bounds may not always hold for each individual case, we further use experiments to validate our analyses in the next section.

%% file: src/experiment.tex
\section{Experimental Results}
In this section, we empirically compare the performance of different treatments on the multiple noisy labels when learning with robust loss functions (CE loss, forward loss correction, and peer loss). 
We consider several treatments including label aggregation methods (majority vote and EM inference) and the label separation method.
Assuming that multiple noisy labels have different weights, EM inference can be used to solve the problem under this assumption by treating the aggregated labels as hidden variables \cite{dawid1979maximum,smyth1994inferring,raykar2010learning,quoc2013evaluation}. In the E-step, the probabilities of the aggregated labels are estimated using the weighted aggregation approach based on the fixed weights of multiple noisy labels. In the M-step, EM inference method re-estimates the weights of multiple noisy labels based on the current aggregated labels. This iteration continues until all aggregated labels remain unchanged.  
As for label separation, we adopted the mini-batch separation method, i.e., each training sample $\x_n$ is assigned with $K$ noisy labels in each batch.
\begin{table*}[!htb]
\caption{The performances of CE/BW/PeerLoss trained on 2 UCI datasets (Breast, and German datasets), with aggregated labels (majority vote, EM inference), and separated labels. We highlight the results with Green (for separation method) and Red (for aggregation methods) if the performance gap is larger than 0.05. ($K$ is the number of labels per training image) 
}
\label{Tab:uci_part1}
\centering{\scalebox{0.79}{
\begin{tabular}{ccccccc||ccccccc}
 \hline \hline
\multicolumn{7}{c}{\emph{UCI-Breast (symmetric)    CE}}& \multicolumn{7}{c}{\emph{UCI-German (symmetric)   CE}}  \\
\hline
 $\epsilon=0.2$&  $K=3$ & $K=5$ & $K=9$ & $K=15$ & $K=25$ & $K=49$ & $\epsilon=0.2$&  $K=3$ & $K=5$ & $K=9$ & $K=15$ & $K=25$ & $K=49$ \\ \hline\hline
 MV & 96.05 & \bad{}96.05 & \bad{}96.49 & \bad{}96.93 & 97.37 & 97.37 & MV   & \bad{}69.00 & \bad{}71.50 & \bad{}71.50 & \bad{}73.50 & \bad{}73.00 & \bad{}73.00  \\
EM  & \bad{}96.93 & \bad{}96.05 &\bad{} 96.49 & \bad{}96.93 & 97.37 & 97.37 & EM & 58.75 & 63.50 & 65.75 & 66.50 & 65.50 & 65.50 \\
Sep  & 96.49 & 95.18 & \good{}96.49 & \good{}96.93 & \good{}97.81 & \good{}98.25 & Sep & \good{}70.00 & 70.75 & 66.00 & 69.75 & 70.75 & 69.25 \\
 \hline
 \hline
 $\epsilon=0.4$&  $K=3$ & $K=5$ & $K=9$ & $K=15$ & $K=25$ & $K=49$ & $\epsilon=0.4$&  $K=3$ & $K=5$ & $K=9$ & $K=15$ & $K=25$ & $K=49$ \\ \hline\hline
 MV& \bad{}96.05 & \bad{}96.49 & 95.18 & 95.18 & 96.49 & \bad{}96.93  & MV & 65.75 & 62.25 & 62.75 & \bad{}68.50 & \bad{}71.75 & \bad{}70.50 \\
EM & \bad{}96.05  & 92.98  & 89.47  & 94.30 & 96.05  & \bad{}96.93 & EM   & 61.00 & 60.00 & 61.50 & 54.00 & 62.00 & 63.25 \\
Sep  & 92.11 & 94.30 & \good{}95.61  & \good{}96.49  & \good{}96.93 &\good{} 96.93  & Sep & \good{}68.25 & \good{}65.50 &\good{} 65.00 & 64.50 & 64.75 & 69.50 \\
 \hline \hline
\multicolumn{7}{c}{\emph{UCI-Breast (symmetric)    BW}}& \multicolumn{7}{c}{\emph{UCI-German (symmetric)   BW}}  \\
\hline
 $\epsilon=0.2$&  $K=3$ & $K=5$ & $K=9$ & $K=15$ & $K=25$ & $K=49$ & $\epsilon=0.2$&  $K=3$ & $K=5$ & $K=9$ & $K=15$ & $K=25$ & $K=49$ \\ \hline\hline
 MV  &  \bad{}95.61  &  \bad{}96.49  & 96.05  &  \bad{}96.93 & 96.93 & 96.93  & MV  & \bad{}72.75 & \bad{}71.50 & \bad{}74.00 & \bad{}75.50 & \bad{}76.50 &\bad{} 76.50 \\
EM  &  \bad{}95.61  & \bad{} 96.49  & 96.05  &  \bad{}96.93 & 96.93 & 96.93 & EM& 62.75 & 61.50 & 59.25 & 64.50 & 62.50 & 62.50 \\
Sep & 95.18 & 93.42  & \good{}96.49  & 96.05  & \good{}97.37 &\good{} 98.25 & Sep & 70.50 & 70.50 & 73.75 & 68.25 & 70.00 & 72.75 \\
 \hline
 \hline
 $\epsilon=0.4$&  $K=3$ & $K=5$ & $K=9$ & $K=15$ & $K=25$ & $K=49$ & $\epsilon=0.4$&  $K=3$ & $K=5$ & $K=9$ & $K=15$ & $K=25$ & $K=49$ \\ \hline\hline
 MV& 89.91  & \bad{}96.05  &  \bad{}94.74 & 94.30 &  \bad{}96.05  & 96.49  & MV &  \bad{}65.25 &  \bad{}69.50 & \bad{} 67.50 &  \bad{}69.50 & \bad{} 70.50 &  \bad{}71.75 \\
EM & 81.14 & 94.30 & 92.11 &  \bad{}94.74 & 92.54  & 96.49  & EM   & 57.75 & 60.25 & 55.25 & 53.50 & 54.00 & 62.25\\
Sep & \good{}91.67 & 93.42  & 94.30 & 89.47  & 92.54  & \good{}97.37  & Sep & 60.25 & 63.50 & 63.00 & 64.25 & 69.00 & 64.75  \\
 \hline \hline
\multicolumn{7}{c}{\emph{UCI-Breast (symmetric)    PeerLoss}}& \multicolumn{7}{c}{\emph{UCI-German (symmetric)   PeerLoss}}  \\
\hline\hline
 $\epsilon=0.2$&  $K=3$ & $K=5$ & $K=9$ & $K=15$ & $K=25$ & $K=49$ & $\epsilon=0.2$&  $K=3$ & $K=5$ & $K=9$ & $K=15$ & $K=25$ & $K=49$ \\ \hline\hline
 MV& \bad{}96.05  & \bad{}96.49  & 96.49  & \bad{}96.93 &\bad{} 96.93 & 96.93  & MV   & \bad{}72.75 & \bad{}71.75 & \bad{}73.00 & \bad{}73.00 &\bad{} 72.50 & 72.50 \\
EM & \bad{}96.05  &\bad{} 96.49  & 96.49  & \bad{}96.93 &\bad{} 96.93 & 96.93 & EM& 62.25 & 64.50 & 63.75 & 64.25 & 62.75 & 62.75 \\
Sep & 94.74 & 94.30 & \good{}96.93 &\good{} 96.93 & \good{}96.93 & \good{}97.81  & Sep & 70.25 & 68.00 & 70.50 & 70.00 & 67.00 & \good{}73.50  \\
 \hline
 \hline
 $\epsilon=0.4$&  $K=3$ & $K=5$ & $K=9$ & $K=15$ & $K=25$ & $K=49$ & $\epsilon=0.4$&  $K=3$ & $K=5$ & $K=9$ & $K=15$ & $K=25$ & $K=49$ \\ \hline\hline
 MV& \bad{}92.11 & \bad{}95.61  &\bad{} 95.18 & 92.54  & \bad{}96.49  & 96.05 & MV & \bad{}69.50 & \bad{}66.25 & \bad{}69.50 & \bad{}68.75 & 69.00 & \bad{}70.00 \\
EM & \bad{}92.11 & 92.11 & 86.40 & 93.86 & 95.61  & \bad{}96.93  & EM  & 62.50 & 61.25 & 64.25 & 57.75 & 59.75 & 65.00\\
Sep  & \good{}92.11 & 94.30 &\good{} 95.18 & \good{}95.18 & 95.61  & 96.05   & Sep & 64.00 & 61.25 & 66.50 & 68.00 &\good{} 69.25 & 69.00 \\
\hline
\end{tabular}}}
\end{table*}

\subsection{Experiment on Synthetic Noisy Datasets}
\begin{figure*}[!htb]
    \centering
    \includegraphics[width = 0.48\textwidth]{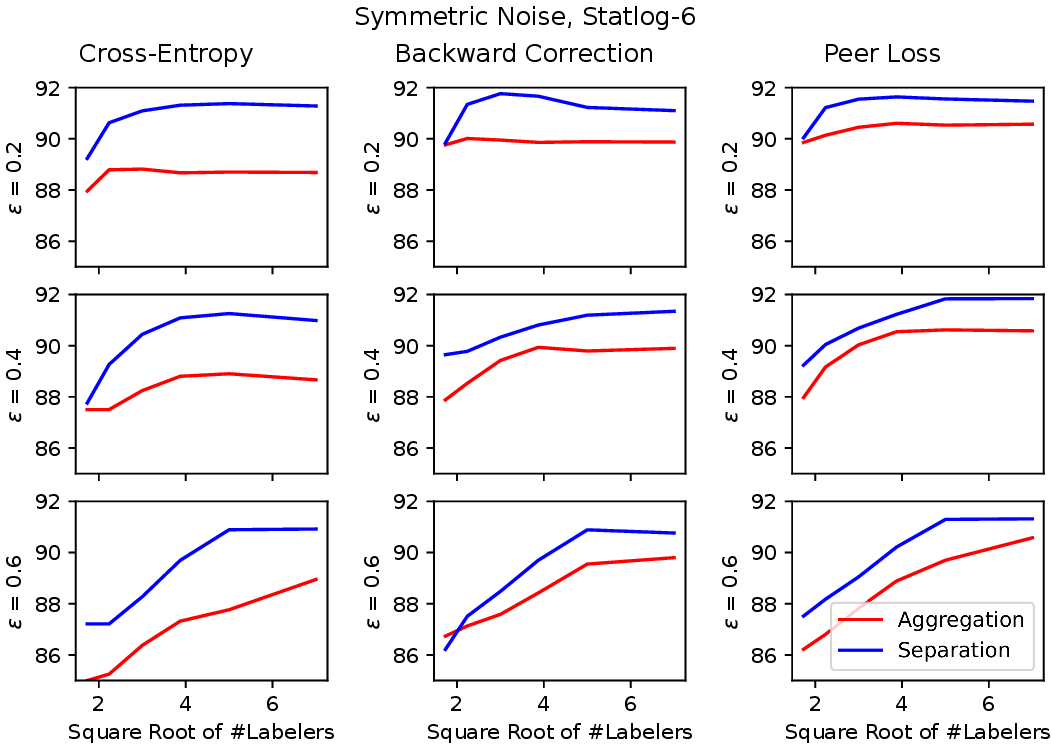}~~
    \includegraphics[width = 0.48\textwidth]{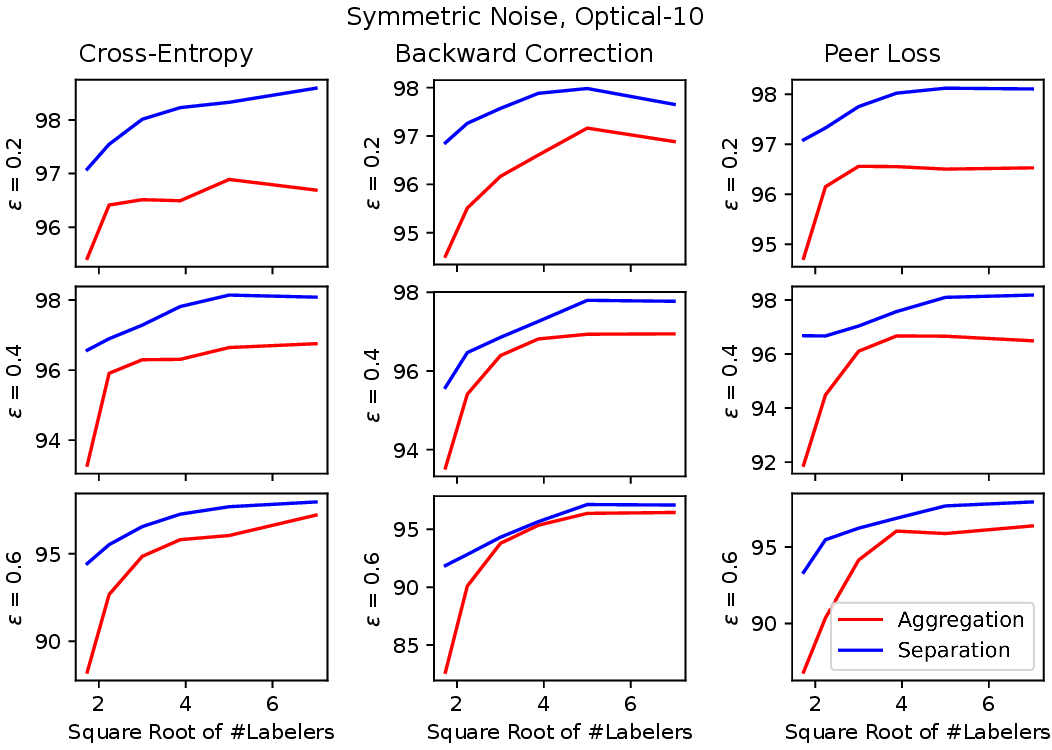}
    \caption{The performances of Cross-Entropy, Backward Loss Correction, and Peer Loss trained on synthetic noisy Statlog-6/Optical-10 aggregated labels (we report the better results between majority vote and EM inference for each $K$, and noise rate $\epsilon$), and separated labels. $X$-axis: the value of the number of labelers $\sqrt{K}$; $Y$ axis: the best test accuracy achieved.}
    \vspace{-0.1in}
    \label{fig:uci}
\end{figure*}

\paragraph{Experimental results on synthetic noisy UCI datasets \cite{Dua:2019}}
We adopt six UCI datasets to empirically compare the performances of label separation and aggregation methods, when learning with CE loss, backward correction \cite{natarajan2013learning,patrini2017making}, and Peer Loss \cite{liu2020peer}. The noisy annotations given by multiple annotators are simulated by \emph{symmetric label noise}, which assumes $T_{i, j}=\frac{\epsilon}{M-1}$ for $j\neq i$ for each annotator, where $\epsilon$ quantifies the overall noise rate of the generated noisy labels.  In Figure \ref{fig:uci}, we adopt two UCI datasets (StatLog: ($M=6$); Optical: ($M=10$)) for illustration. From the results in Figure~\ref{fig:uci}, it is quite clear that: the \emph{label separation method outperforms both aggregation methods (majority-vote and EM inference) consistently, and is considered to be more beneficial on such small scale datasets}. Results on additional datasets and more details are deferred to the Appendix.

\paragraph{Experimental results on synthetic noisy CIFAR-10 dataset \cite{krizhevsky2009learning}}
On CIFAR-10 dataset, we consider two types of simulation for the separate noisy labels: \emph{symmetric label noise} model and \emph{instance-dependent label noise} \cite{cheng2020learning,zhu2021clusterability}, where $\epsilon$ is the average noise rate and different labelers follow different instance-dependent noise transition matrices. For a fair comparison, we adopt the ResNet-34 model~\cite{he2016deep}, the same training procedure and batch-size for all considered treatments on the separate noisy labels.

\begin{figure*}[!htb]
    \centering
    \includegraphics[width = 0.48\textwidth]{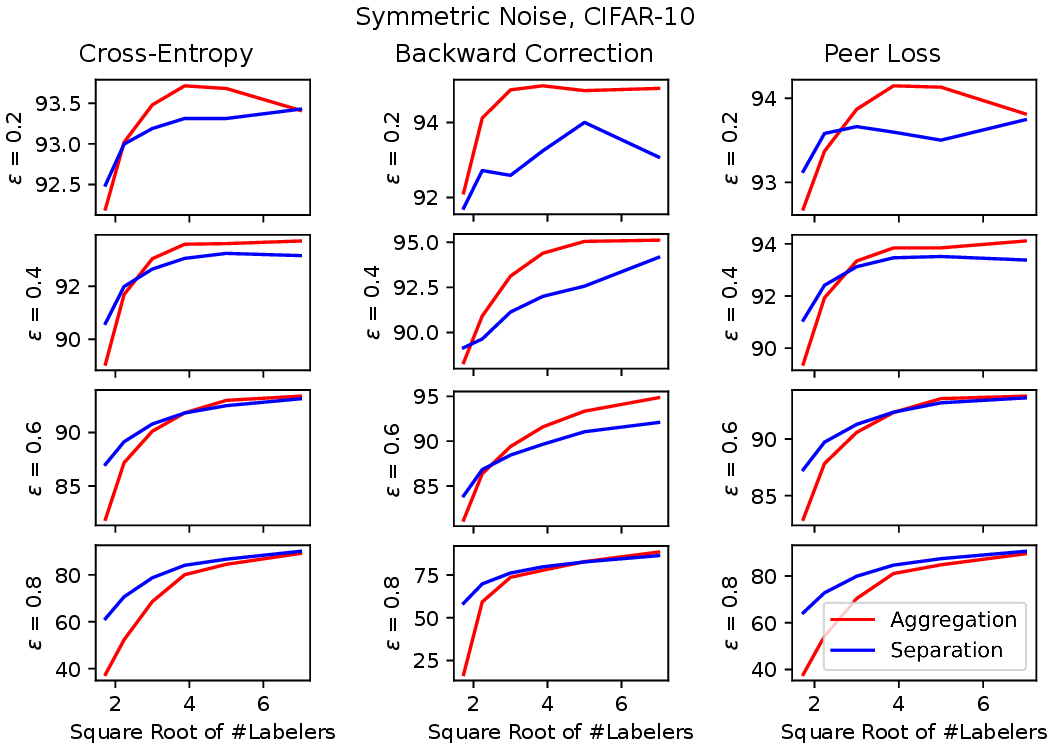}~~
    \includegraphics[width = 0.48\textwidth]{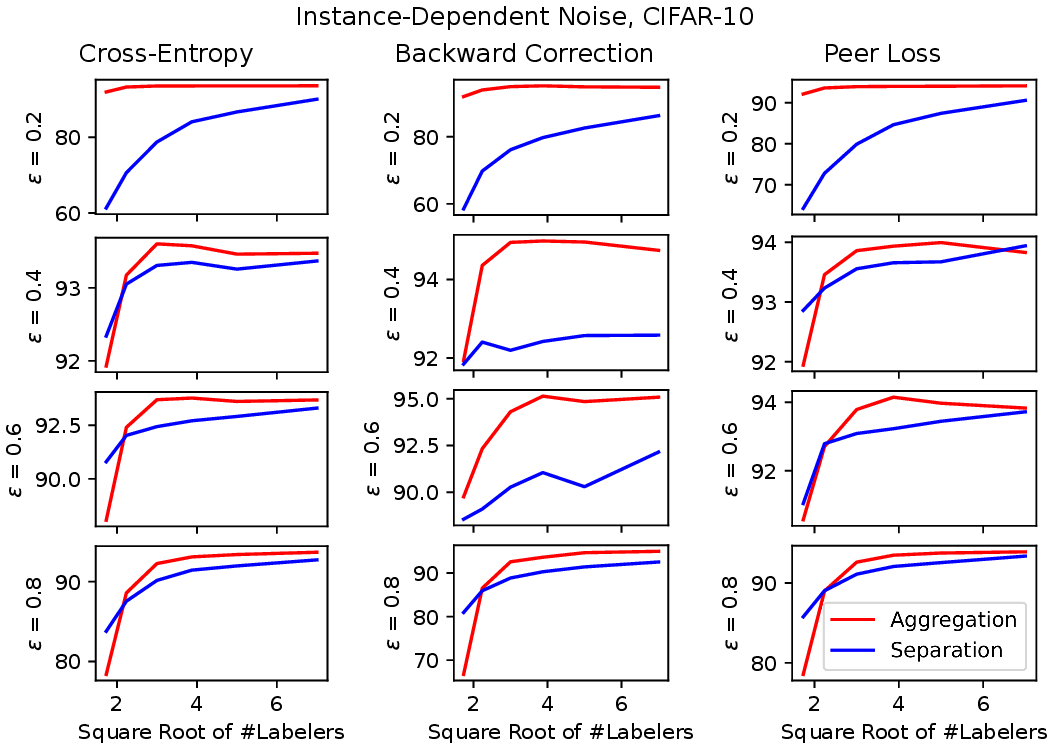}
    \caption{The performances of Cross-Entropy, Backward Loss Correction, and Peer Loss trained on  synthetic noisy CIFAR-10 aggregated labels (we report the better results between majority vote, EM inference for each $K$, and noise rate $\epsilon$), and separated labels. $X$-axis: the value of $\sqrt{K}$ where $K$ denotes the number of labels per training example; $Y$ axis: the best achieved test accuracy. }
    \vspace{-0.1in}
    \label{fig:c10}
\end{figure*}

Figure \ref{fig:c10} shares the following insights regarding the preference of the treatments: in the low noise regime or when $K$ is large, aggregating separate noisy labels significantly reduces the noise rates and aggregation methods tend out to have a better performance; while in the high noise regime or when $K$ is small, the performances of separation methods tend out to be more promising. With the increasing of $K$ or $\epsilon$, we can observe a preference transition from label separation to label aggregation methods.

\subsection{ Empirical Verification of the Theoretical Bounds}

To verify the comparisons of bias proxies (i.e., Theorem \ref{tm:bounds_ce}) through an empirical perspective, we adopt two binary classification UCI datasets for demonstration: Breast and German datasets, as shown in Table \ref{Tab:uci_part1}. Clearly, on these two binary classification tasks, label aggregation methods tend to outperform label separation, and we attribute this phenomenon to the fact that ``denoising effect of label aggregation is more significant in the binary case''.

\begin{table}[!htb]
\caption{Empirical verification of Theorem \ref{tm:bounds_ce} on Breast \& German UCI datasets.}
\label{tab: verify_theorem1}
\centering
{\scalebox{1.0}{\begin{tabular}{|c| c| c| c| c|  }
\hline
\textbf{Dataset}  & $\rho_i^{\circ}$  & $p_0$ & $N$ & $(1-\delta, S_K)$\\ \hline
Breast & $0.2$  & 0.3726 & 569 & $(0.62
, \{K>49\})$ \\
Breast & $0.4$   & 0.3726 & 569 & $(0.62,  \{K>49\})$ \\\hline
German & $0.2$   & 0.3 & 1000 &$(0.98,  \{K>15\})$\\
German & $0.4
$   & 0.3 & 1000 &$(0.98,  \{K>23\})$\\
 \hline
\end{tabular}}}
\end{table}

For Theorem \ref{tm:bounds_ce} (CE loss), the condition requires $\alpha_K/\left(1-(\eta_K^{\circ})^{-\frac{1}{2}}\right)$, where $\alpha=(\rho_0^{\circ}p_0+\rho_1^{\circ}p_1)-(\rho_0^{\bullet}p_0+\rho_1^{\bullet}p_1)$, $\gamma=\sqrt{\log(1/\delta)/2N}$. For two binary UCI datasets (Breast \& German), the information could be summarized in Table \ref{tab: verify_theorem1}, where the column $(1-\delta, S_K)$ means: when the number of annotators belongs to the set $S_K$, the label separation method is likely to under-perform label aggregation (i.e., majority vote) with probability at least $1-\delta$. For example, in the last row of Table \ref{tab: verify_theorem1}, when training on UCI German dataset with CE loss under noise rate $0.4$ (the noise rate of separate noisy labels), Theorem \ref{tm:bounds_ce} reveals that with probability at least 0.98, label aggregation (with majority vote) is better than label separation when $K>23$, which aligns well with our empirical observations (label separation is better only when $K<15$). 

\subsection{Experiments on realistic noisy datasets}
Note that in real-world scenarios, the label-noise pattern may differ due to the expertise of each human annotator. We further compare the different treatments on two realistic noisy datasets: CIFAR-10N \cite{wei2021learning}, and CIFAR-10H \cite{peterson2019human}. CIFAR-10N provides each CIFAR-10 train image with 3 independent human annotations, while CIFAR-10H gives $\approx 50$ annotations for each CIFAR-10 test image.

In Table \ref{tab:real_k3}, we repeat the reproduce of three robust loss functions with three different treatments on the separate noisy labels. We report the best achieved test accuracy for Cross-Entropy/Backward Correction/Peer Loss methods when learning with label aggregation methods (majority-vote and EM inference) and the separation method (soft-label). We observe that the separation method tends to have a better performance than aggregation ones. This may be attributed to the relative high noise rate ($\epsilon\approx 0.18$) in CIFAR-N and the insufficient amount of labelers ($K=3$).  Note that since the noise level in CIFAR-10H is low ($\epsilon\approx 0.07$ wrong labels), label aggregation methods can infer higher quality labels, and thus, result in a better performance than separation methods (Red colored cells in Table \ref{tab:real_k3} and \ref{Tab:cifarh}).
\begin{table}[!htb]
 \caption{Experimental results on CIFAR-10N and CIFAR-10H dataset with $K=3$. We highlight the results with Green (for separation method) and Red (for aggregation methods) if the performance gap is large than $0.05$. 
}\label{tab:real_k3}
\centering
   {\scalebox{1.0}{ \begin{tabular}{c|ccc}
\hline
 \textbf{CIFAR-10N} ($\epsilon\approx 0.18$)& \textbf{CE} & \textbf{BW} & \textbf{PL}  \\ \hline\hline
 Majority-Vote & 89.52 & \bad{89.23} & 89.84 \\
 EM-Inference & 89.19 &  88.88  & 88.92 \\
  Separation & \good{89.77}& \good{89.20}  & \good{89.97} \\
 \hline \hline
 \textbf{CIFAR-10H} ($\epsilon\approx 0.09$)& \textbf{CE} & \textbf{BW} & \textbf{PL} \\ \hline\hline
 Majority-Vote &  \bad{80.86} &  \bad{82.72} &  \bad{82.11}\\
 EM-Inference & 80.81  & 82.43 & 81.73\\
  Separation & 76.75 & 79.07 & 78.08\\
  \hline
\end{tabular}}}
\end{table}
\begin{table}[!htb]
\centering
        \caption{Experimental results on CIFAR10-H with $K\geq 5$. We highlight the results with Green (for separation method) and Red (for aggregation methods) if the performance gap is large than $0.05$. 
}
\label{Tab:cifarh}
{\scalebox{0.9}{\begin{tabular}{c|c|c|c|c|c}
\hline
 \textbf{CE}& $K=5$ & $K=9$ & $K=15$ & $K=25$ & $K=49$\\ \hline\hline
 Majority-Vote & 80.69& 80.73  &\bad{81.37} & \bad{81.79} &81.66 \\
 EM-Inference   &\bad{80.97} & \bad{80.96}& 81.24& 81.01& \bad{81.68}\\
 Separation   & 79.65 & 80.91&81.07 & 80.78 & 80.81\\
 \hline
 \textbf{BW} & $K=5$ & $K=9$ & $K=15$ & $K=25$ & $K=49$\\ \hline\hline
 Majority-Vote & \bad{82.51} &  \bad{82.75}& \bad{83.27}  & \bad{83.59} & \bad{83.68} \\
 EM-Inference  & 82.30 &  82.68& 82.74  & 82.89 & 83.08 \\
 Separation   & 82.14 &  82.48 & 81.92  & 81.72 & 81.69 \\
 \hline
 \textbf{PL} & $K=5$ & $K=9$ & $K=15$ & $K=25$ & $K=49$\\ \hline\hline
 Majority-Vote & 81.84 &81.85 & 82.39 & \bad{82.98}& \bad{82.83}\\
 EM-Inference  & \bad{81.89} & \bad{82.30} & \bad{82.53}& 82.86& 82.73\\
 Separation   & 80.25 &81.89&81.00 & 80.71 & 80.89\\
 \hline
\end{tabular}}}
\end{table}

\subsection{Hypothesis Testing}

We adopt the paired t-test to show which treatment on the separate noisy labels is better, under certain conditions. In Table \ref{tab:t_test}, we report the statistic and $p$-value given by the hypothesis testing results. The column ``Methods'' indicate the two methods we want to compare (A \& B). Positive statistics means that A is better than B in the metric of test accuracy. Given a specific setting, denote by $\text{Acc}_{\text{method}}$ as the list of test accuracy that belongs to this setting (i.e., CIFAR-10N, $K=3$), including CE, BW, PL loss functions, the basic hypothesis could be summarized as below:
\squishlist
    \item \textbf{Null hypothesis}: there exists zero mean difference between (1) $\text{Acc}_{\text{MV}}$ and $\text{Acc}_{\text{EM}}$; or (2) $\text{Acc}_{\text{MV}}$ and $\text{Acc}_{\text{Sep}}$; or (3) $\text{Acc}_{\text{EM}}$ and $\text{Acc}_{\text{Sep}}$; 
    \item \textbf{Alternative hypothesis}: there exists non-zero mean difference between (1) $\text{Acc}_{\text{MV}}$ and $\text{Acc}_{\text{EM}}$; or (2) $\text{Acc}_{\text{MV}}$ and $\text{Acc}_{\text{Sep}}$; or (3) $\text{Acc}_{\text{EM}}$ and $\text{Acc}_{\text{Sep}}$.
\squishend
To clarify, the three cases in the above hypothesis are tested independently.
For test accuracy comparisons of CIFAR-10N in Table \ref{tab:real_k3}, the setting of hypothesis test is $K=3$ and the label noise rate is relatively high (18\%). All $p$-values are larger than $0.05$, indicating that we should reject the null hypothesis, and we can conclude that the performance of these three methods on CIFAR-10N (high noise, small $K$) satisfies: EM<MV<Sep.

For CIFAR-10H in Table \ref{tab:real_k3} and \ref{Tab:cifarh}, all the label noise rate is relatively low. We consider two scenarios ($K<15$: the number of annotators is small; $K\geq 15$: the number of annotators is large). $p$-values among MV and EM are always large, which mean that the denoising effect of the advanced label aggregation method (EM) is negligible under CIFAR-10H dataset. However, $p$-values of remaining settings are larger than $0.05$, indicating that we should reject the null hypothesis, and we can conclude that the performance of these 3 methods on CIFAR-10H (low noise, small/large $K$) satisfies: EM/MV > Sep.

\begin{table}[!htb]\centering
\caption{Hypothesis testing results of the comparisons between label aggregation methods and the label separation method on realistic noisy datasets.}
\label{tab:t_test}
{\scalebox{0.9}{\begin{tabular}{|c| c |cc|}
\hline
Setting  & Methods & Statistic & $p$-value \\ \hline
CIFAR-10N ($K= 3$, high noise) & MV \& EM & 2.650 & 0.057     \\ 
CIFAR-10N ($K= 3$, high noise) & MV \& Sep & -0.401 & 0.708    \\ 
CIFAR-10N ($K= 3$, high noise) & EM \& Sep & -2.596 & 0.060    \\ \hline
CIFAR-10H ($K<15$, low noise) & MV \& EM & -0.003 & 0.998    \\ CIFAR-10H ($K<15$, low noise) & MV \& Sep & 2.336 & 0.033   \\ CIFAR-10H ($K<15$, low noise) & EM \& Sep & 2.390  & 0.030   \\\hline
CIFAR-10H ($K\geq 15$, low noise) & MV \& EM & 0.805  &   0.433   \\ CIFAR-10H ($K\geq 15$, low noise) & MV \& Sep & 4.426 &  0.000 \\ CIFAR-10H ($K\geq 15$, low noise) & EM \& Sep &  3.727  &  0.002  \\
 \hline
\end{tabular}}}
\end{table}

%% file: src/appendices.tex
\section{Proof Sketch of Core Theorems}

We briefly introduce the proof sketch of Lemma \ref{lm:bound} because it sets up the foundation for the analyses on Backward Loss Correction and it covers the proofs of the standard $\ell$ loss in Section \ref{sec:ell_theory} as a special case.
\subsection{Proof of Lemma \ref{lm:bound}}

\begin{proof}
Our proof can be divided into four steps as follows.
\paragraph{Step 1: Apply Hoeffding's inequality for each group.}
We divide the noisy train samples $\{(x_n, \ny_{n,k}^{\sep})\}_{n\in[N]}$ into $K$ groups, for $k\in[K]$, i.e., $\{(x_n, \ny_{n,1}^{\sep})\}_{n\in[N]}$, $\cdots$, $\{(x_n, \ny_{n,K}^{\sep})\}_{n\in[N]}$. Note within each group, e.g., group $\{(x_n, \ny_{n,1}^{\sep})\}_{n\in[N]}$, all the $N$ training samples are i.i.d. Additionally, training samples between any two different groups are also i.i.d. given feature set $\{x_n\}_{n\in[N]}$. Thus, with one group $\{(x_n, \ny_{n,1})\}_{n\in[N]}$, w.p. $1-\delta_0$, we have
\[\left| \hat{R}_{\BR_{\slc}^{\sep}|\text{Group-1}}(f)-R_{\BR_{\slc}^{\sep}}(f)\right|
\leq \left(\overline{\BR_{\slc}^{\sep}}-\underline{\BR_{\slc}^{\sep}}\right) \cdot \sqrt{\frac{\log(1/\delta_0)}{2N}}, \forall f.
\]
where we have $\overline{\BR_{\slc}^{\sep}}-\underline{\BR_{\slc}^{\sep}}:= L_{\slc0}^{\sep} =\frac{(1+|\rho^{\sep}_0-\rho^{\sep}_1|)}{1-\rho^{\sep}_0-\rho^{\sep}_1}$.
\paragraph{Step 2: Adopt the union bound for all groups.}
Applying the above technique on the other groups and by the union bound, we know that w.p. at least $1-K\delta_0$, $ \forall k\in[K]$, each $\hat R_{\BR_{\slc}^{\sep}|\text{Group-k}}(f), k\in[K]$ can be seen as a random variable within range:  \[\left[R_{\BR_{\slc}^{\sep}}(f) - L_{\slc0}^{\sep}\cdot\sqrt{\frac{\log(1/\delta_0)}{2N}},R_{\BR_{\slc}^{\sep}}(f) +L_{\slc0}^{\sep}\cdot \sqrt{\frac{\log(1/\delta_0)}{2N}}\right].\] The randomness is from noisy labels $\tilde y_{n,k}$.
\paragraph{Step 3: Hoeffding inequality for $\hat R_{\BR_{\slc}^{\sep}|\text{Group-k}}(f), k\in[K]$}
These $K$ random variables are i.i.d. when the feature set is fixed. By Hoeffding's inequality, w.p. at least $1-K\delta_0-\delta_1$, $\forall f$, we have
\begin{align*}
&\left| \hat R_{\BR_{\slc}^{\sep}}(f)-R_{\BR_{\slc}^{\sep}}(f)\right|\leq   L_{\slc0}^{\sep}\cdot \sqrt{\frac{\log(1/\delta_1)\log(1/\delta_0)}{NK}}.
\end{align*}

\paragraph{Step 4: Rademacher bound on the maximal deviation}
For $\delta_0=\delta_1=\frac{\delta}{K+1}$, with the Rademacher bound on the maximal deviation between risks and empirical ones, for $\f\in \mathcal{F}$ and the separation method, with probability at least $1-\delta$, we have:
\begin{align*}
    &\max_{f\in \mathcal{F}}\left|\hr_{\ell_{\slc}, \nDe^{\sep}}(f)-R_{\ell_{\slc}, \nD^{\sep}}(f)\right|\leq 2 \Rk^{\circ}(\ell_{\slc} \circ \mathcal{F})+L_{\slc0}^{\sep}\cdot (\overline{\ell}- \underline{\ell}) \cdot\log\left(\frac{K+1}{\delta}\right)\cdot\sqrt{\frac{1}{NK}},\\
    & \max_{f\in \mathcal{F}}\left|\hr_{\ell_{\slc}, \nDe^{\agg}}(f)-R_{\ell_{\slc}, \nD^{\agg}}(f)\right|\leq  2 \Rk^{\agg}(\ell_{\slc} \circ \mathcal{F})+L_{\slc0}^{\agg}\cdot (\overline{\ell}- \underline{\ell}) \cdot\sqrt{\frac{\log(1/\delta)}{2N}},
\end{align*}
where we define $\overline{\ell}, \underline{\ell}$ as the upper and lower bound of loss function $\ell$ respectively, and $\Rk^{\uni}(\ell_{\slc} \circ \mathcal{F})$ is the Rademacher complexity.

\paragraph{Step 5: Adopt the Lipshitz composition property of Rademacher averages.}
If $\ell$ is $L-$Lipshitz, then for separation and aggregation methods, $\ell_{\slc}$ is $L_{\slc}^{\uni}$ Lipshitz with $L_{\slc}^{\uni}=\frac{(1+|\rho^{\uni}_0-\rho^{\uni}_1|)L}{1-\rho^{\uni}_0-\rho^{\uni}_1}$.
\paragraph{Step 6: Triangle inequality}

Bound with the triangle inequality:
\begin{align*}
    &R_{\ell, \D}(\hf^{\uni}_{\slc})-R_{\ell, \D}(f^*)=\underline{R_{\ell_{\slc}, \nD^{\uni}}(\hf^{\uni}_{\slc})-R_{\ell_{\slc}, \nD^{\uni}}(f^*)}\\
    =&\underline{R_{\ell_{\slc}, \nD^{\uni}} (\hf^{\uni}_{\slc})}-\hr_{\ell_{\slc}, \nDe^{\uni}} (\hf^{\uni}_{\slc})+\hr_{\ell_{\slc}, \nDe^{\uni}} (f^*)-\underline{R_{\ell_{\slc}, \nD^{\uni}} (f^*)} +\hr_{\ell_{\slc}, \nDe^{\uni}} (\hf^{\uni}_{\slc})-\hr_{\ell_{\slc}, \nDe^{\uni}}(f^*)\\
    \leq & 0+2\max_{f\in \mathcal{F}}|\hr_{\ell_{\slc}, \nDe^{\uni}}(f)-R_{\ell_{\slc}, \nD^{\uni}}(f)|.
\end{align*}
Conclusions could be derived then.
\end{proof}

%% file: src/proofs.tex
\section{Full Proofs}
In this section, we briefly introduce all omitted proofs in the main paper.

We firstly give the proof of Lemma \ref{lm:bound} because it is beneficial for the proofs in Section \ref{sec:ell_theory}.
\subsection{Proof of Lemma \ref{lm:bound}}

\begin{proof}

To apply Hoeffding's inequality on the dataset of the separation method, we divide the noisy train samples $\{(x_n, \ny_{n,k}^{\sep})\}_{n\in[N]}$ into $K$ groups, for $k\in[K]$, i.e., $\{(x_n, \ny_{n,1}^{\sep})\}_{n\in[N]}$, $\cdots$, $\{(x_n, \ny_{n,K}^{\sep})\}_{n\in[N]}$. Note within each group, e.g., group $\{(x_n, \ny_{n,1}^{\sep})\}_{n\in[N]}$, all the $N$ training samples are i.i.d. Additionally, training samples between any two different groups are also i.i.d. given feature set $\{x_n\}_{n\in[N]}$. Thus, with one group $\{(x_n, \ny_{n,1})\}_{n\in[N]}$, w.p. $1-\delta_0$, we have
\[\left| \hat{R}_{\BR_{\slc}^{\sep}|\text{Group-1}}(f)-R_{\BR_{\slc}^{\sep}}(f)\right|\leq \left(\overline{\BR_{\slc}^{\sep}}-\underline{\BR_{\slc}^{\sep}}\right) \cdot \sqrt{\frac{\log(1/\delta_0)}{2N}}, \forall f.
\]
Note that:
\begin{align*}
    (T^{\uni})^{-1}=\dfrac{1}{1-\rho^{\uni}_0-\rho^{\uni}_1}\Big(\begin{matrix}
    1-\rho^{\uni}_{1} & -\rho^{\uni}_{0} \\
    -\rho^{\uni}_{1} & 1-\rho^{\uni}_{0}
\end{matrix}\Big), \quad \text{for } \uni\in\{\sep, \agg\},
\end{align*}
we have:
\begin{align*}
    \overline{\BR_{\slc}^{\sep}}-\underline{\BR_{\slc}^{\sep}}:=  L_{\slc0}^{\sep} =\frac{(1+|\rho^{\sep}_0-\rho^{\sep}_1|)}{1-\rho^{\sep}_0-\rho^{\sep}_1}.
\end{align*}

Applying the above technique on the other groups and by the union bound, we know that w.p. at least $1-K\delta_0$, $ \forall k\in[K]$,
\begin{align*}
&\hat R_{\BR_{\slc}^{\sep}|\text{Group-k}}(f) \in  \left[R_{\BR_{\slc}^{\sep}}(f) - L_{\slc0}^{\sep}\cdot\sqrt{\frac{\log(1/\delta_0)}{2N}},R_{\BR_{\slc}^{\sep}}(f) +L_{\slc0}^{\sep}\cdot \sqrt{\frac{\log(1/\delta_0)}{2N}}\right].
\end{align*}
Each $\hat R_{\BR_{\slc}^{\sep}|\text{Group-k}}(f), k\in[K]$ can be seen as a random variable within range: \[\left[R_{\BR_{\slc}^{\sep}}(f) - L_{\slc0}^{\sep}\cdot\sqrt{\frac{\log(1/\delta_0)}{2N}},R_{\BR_{\slc}^{\sep}}(f) +L_{\slc0}^{\sep}\cdot \sqrt{\frac{\log(1/\delta_0)}{2N}}\right].\] The randomness is from noisy labels $\tilde y_{n,k}$.
Recall that the samples between different groups are i.i.d. given $\{x_n\}_{n\in[N]}$. Then the above $K$ random variables are i.i.d. when the feature set is fixed. By Hoeffding's inequality, w.p. at least $1-K\delta_0-\delta_1$, $\forall f$, we have
\begin{align*}
\left| \hat R_{\BR_{\slc}^{\sep}}(f)-R_{\BR_{\slc}^{\sep}}(f)\right| &\leq  2 \cdot L_{\slc0}^{\sep} \cdot  \sqrt{\frac{\log(1/\delta_0)}{2N}} \cdot \sqrt{\frac{\log(1/\delta_1)}{2K}} = L_{\slc0}^{\sep}\cdot \sqrt{\frac{\log(1/\delta_1)\log(1/\delta_0)}{NK}}.
\end{align*}

For $\delta_0=\delta_1=\frac{\delta}{K+1}$, with the Rademacher bound on the maximal deviation between risks and empirical ones, for $\f\in \mathcal{F}$ and the separation method, with probability at least $1-\delta$, we have:
\begin{align*}
    \max_{f\in \mathcal{F}}\left|\hr_{\ell_{\slc}, \nDe^{\sep}}(f)-R_{\ell_{\slc}, \nD^{\sep}}(f)\right|&\leq 2 \Rk^{\circ}(\ell_{\slc} \circ \mathcal{F})+L_{\slc0}^{\sep}\cdot (\overline{\ell}- \underline{\ell}) \cdot\log\left(\frac{K+1}{\delta}\right)\cdot\sqrt{\frac{1}{NK}},\\
     \max_{f\in \mathcal{F}}\left|\hr_{\ell_{\slc}, \nDe^{\agg}}(f)-R_{\ell_{\slc}, \nD^{\agg}}(f)\right|&\leq 2 \Rk^{\agg}(\ell_{\slc} \circ \mathcal{F})+\left(\overline{\ell_{\slc}^{\agg}}-\underline{\ell_{\slc}^{\agg}}\right)\cdot\sqrt{\frac{\log(1/\delta)}{2N}}\\
    &= 2 \Rk^{\agg}(\ell_{\slc} \circ \mathcal{F})+L_{\slc0}^{\agg}\cdot (\overline{\ell}- \underline{\ell}) \cdot\sqrt{\frac{\log(1/\delta)}{2N}},
\end{align*}
where we define $\overline{\ell}, \underline{\ell}$ as the upper and lower bound of loss function $\ell$ respectively, and:
\begin{align*}
    \Rk^{\sep}(\ell_{\slc} \circ \mathcal{F})&:=\mathbb{E}_{\x_i, \ny_{i,1}^{\sep}, ..., \ny_{i,K}^{\sep},\epsilon_i}\left[\sup_{f\in\mathcal{F}}\frac{1}{NK}\sum_{i=1}^N\sum_{j=1}^K\epsilon_i \ell_{\slc}(f(\x_i), \ny_{i, j}^{\sep})\right]\\
    &\leq \frac{1}{K}\sum_{j=1}^K \mathbb{E}_{\x_i, \ny_{i,j}^{\sep}, \epsilon_i}\left[\sup_{f\in\mathcal{F}}\frac{1}{N}\sum_{i=1}^N\epsilon_i \ell_{\slc}(f(\x_i), \ny_{i, j}^{\sep})\right],
\end{align*}
\begin{align*}
    \Rk^{\agg}(\ell_{\slc} \circ \mathcal{F})&:=\mathbb{E}_{\x_i, \ny^{\agg},\epsilon_i}\left[\sup_{f\in\mathcal{F}}\frac{1}{N}\epsilon_i \ell_{\slc}(f(\x_i), \ny^{\agg})\right].
\end{align*}
Note that we assume the noisy labels given by the $K$ labelers follow the same noise transition matrix, if $\ell$ is $L-$Lipshitz, then for separation and aggregation methods, $\ell_{\slc}$ is $L_{\slc}^{\uni}$ Lipshitz for $\uni\in\{\sep, \agg\}$ respectively, where $L_{\slc}^{\uni}=\frac{(1+|\rho^{\uni}_0-\rho^{\uni}_1|)L}{1-\rho^{\uni}_0-\rho^{\uni}_1}\leq \frac{2L}{1-\rho^{\uni}_0-\rho^{\uni}_1}$.
By the Lipshitz composition property of Rademacher averages, we have: $\Rk^{\uni}(\ell_{\slc}\circ \mathcal{F})\leq L_{\slc}^{\uni} \cdot \Rk(\mathcal{F})$.
Thus, we have:
\begin{align}
&\max_{f\in \mathcal{F}}|\hr_{\ell_{\slc}, \nDe^{\sep}}(f)-R_{\ell_{\slc}, \nD^{\sep}}(f)|\leq 2L_{\slc}^{\sep} \Rk(\mathcal{F})+\frac{(1+|\rho^{\sep}_0-\rho^{\sep}_1|)\cdot (\overline{\ell}- \underline{\ell})}{1-\rho^{\sep}_0-\rho^{\sep}_1}\cdot\log(\frac{K+1}{\delta})\cdot\sqrt{\frac{1}{NK}},
\end{align}
\begin{align}
&\max_{f\in \mathcal{F}}|\hr_{\ell_{\slc}, \nDe^{\agg}}(f)-R_{\ell_{\slc}, \nD^{\agg}}(f)|\leq 2L_{\slc}^{\agg} \Rk(\mathcal{F})+\frac{(1+|\rho^{\agg}_0-\rho^{\agg}_1|)\cdot (\overline{\ell}- \underline{\ell})}{1-\rho^{\agg}_0-\rho^{\agg}_1}\cdot\sqrt{\frac{\log(1/\delta)}{2N}}.\end{align}

Assume $f^*\slc \min_{f\in \mathcal{F}}R_{\ell, \D}(f)$, for separation methods, we further have:
\begin{align*}
    &R_{\ell, \D}(\hf^{\sep}_{\slc})-R_{\ell, \D}(f^*)=\underline{R_{\ell_{\slc}, \nD^{\sep}}(\hf^{\sep}_{\slc})-R_{\ell_{\slc}, \nD^{\sep}}(f^*)}\\
    &=\underline{R_{\ell_{\slc}, \nD^{\sep}} (\hf^{\sep}_{\slc})}-\hr_{\ell_{\slc}, \nDe^{\sep}} (\hf^{\sep}_{\slc})+\hr_{\ell_{\slc}, \nDe^{\sep}} (f^*)-\underline{R_{\ell_{\slc}, \nD^{\sep}} (f^*)}+\hr_{\ell_{\slc}, \nDe^{\sep}} (\hf^{\sep}_{\slc})-\hr_{\ell_{\slc}, \nDe^{\sep}}(f^*)\\
    &\leq 0+2\max_{f\in \mathcal{F}}|\hr_{\ell_{\slc}, \nDe^{\sep}}(f)-R_{\ell_{\slc}, \nD^{\sep}}(f)|\\
    &\leq 4L_{\slc}^{\sep} \Rk(\mathcal{F})+2L_{\slc}^{\sep}\cdot (\overline{\ell}- \underline{\ell}) \cdot\log(\frac{K+1}{\delta})\cdot\sqrt{\frac{1}{NK}}.
\end{align*}
Similarly, for aggregation methods, we have:
\begin{align*}
    &R_{\ell, \D}(\hf^{\agg}_{\slc})-R_{\ell, \D}(f^*)=\underline{R_{\ell_{\slc}, \nD^{\agg}}(\hf^{\agg}_{\slc})-R_{\ell_{\slc}, \nD^{\agg}}(f^*)}\\
    &=\underline{R_{\ell_{\slc}, \nD^{\agg}}(\hf^{\agg}_{\slc})}-\hr_{\ell_{\slc}, \nDe^{\agg}}(\hf^{\agg}_{\slc})+\hr_{\ell_{\slc}, \nDe^{\agg}}(f^*)-\underline{R_{\ell_{\slc}, \nD^{\agg}}(f^*)}+\hr_{\ell_{\slc}, \nDe^{\agg}}(\hf^{\agg}_{\slc})-\hr_{\ell_{\slc}, \nDe^{\agg}}(f^*)\\
    &\leq 0+2\max_{f\in \mathcal{F}}|\hr_{\ell_{\slc}, \nDe^{\agg}}(f)-R_{\ell_{\slc}, \nD^{\agg}}(f)|\\
    &\leq 4L_{\slc}^{\agg} \Rk(\mathcal{F})+2L_{\slc}^{\agg}\cdot (\overline{\ell}- \underline{\ell})\cdot\sqrt{\frac{\log(1/\delta)}{2N}}.
\end{align*}
Note that $\eta^{\sep}_K=\frac{K\cdot \log(\frac{1}{\delta})}{2\left(\log(\frac{K+1}{\delta})\right)^2}$ and $\eta^{\agg}_K\equiv 1$, we then have:
$$R_{\ell, \D}(\hf_{\slc}^{\uni})-R_{\ell, \D}(f^*) 
\leq  \underbrace{4L_{\slc}^{\uni} \Rk(\mathcal{F})+\frac{L_{\slc}^{\uni} \cdot (\overline{\ell}- \underline{\ell})}{L}\cdot\sqrt{\frac{2\log(1/\delta)}{\eta^{\uni}_K N}}}_{\text{Defined as: } {\pb^{\uni}}_{\slc}}.$$
\end{proof}

\subsection{Proof of Theorem \ref{tm:bounds}}
\begin{proof}
The proof is straightforward if we proceed the proof of Lemma \ref{lm:bound} with below discussions.
With the knowledge of noise rates for both methods, remember that $L_{\slc}^{\uni}=\frac{(1+|\rho^{\uni}_0-\rho^{\uni}_1|)L}{1-\rho^{\uni}_0-\rho^{\uni}_1}$, we have:
{
\begin{align*}
    {\pb^{\sep}}_{\slc}<{\pb^{\agg}}_{\slc} &\Longrightarrow \quad 2L_{\slc}^{\sep} \Rk(\mathcal{F})+\frac{L_{\slc}^{\sep}}{L}\cdot (\overline{\ell}-\underline{\ell}) \cdot \log(\frac{K+1}{\delta})\cdot\sqrt{\frac{1}{NK}}\\
    &~~~\qquad <2L_{\slc}^{\agg} \Rk(\mathcal{F})+\frac{L_{\slc}^{\agg}}{L}\cdot (\overline{\ell}-\underline{\ell})\cdot \sqrt{\frac{\log(1/\delta)}{2N}}
    \\
    &\Longrightarrow \quad 2\cdot \frac{L_{\slc}^{\sep}-L_{\slc}^{\agg}}{\overline{\ell}-\underline{\ell}}\cdot L \cdot \Rk(\mathcal{F})< L_{\slc}^{\agg}\cdot\sqrt{\frac{\log(1/\delta)}{2N}}-L_{\slc}^{\sep}\cdot\log(\frac{K+1}{\delta})\cdot\sqrt{\frac{1}{NK}}\\
    &\Longrightarrow \quad 2\cdot \frac{L_{\slc}^{\sep}-L_{\slc}^{\agg}}{\overline{\ell}-\underline{\ell}}\cdot L \cdot \Rk(\mathcal{F}) <\left(L_{\slc}^{\agg}-L_{\slc}^{\sep}\cdot \sqrt{\frac{1}{\eta^{\sep}_K}}\right) \sqrt{\frac{\log(1/\delta)}{2N}}\\
    &\Longrightarrow \quad 2\sqrt{\frac{2N}{\log(1/\delta)}}\frac{L_{\slc}^{\sep}-L_{\slc}^{\agg}}{\overline{\ell}-\underline{\ell}}\cdot L \cdot\Rk(\mathcal{F})<L_{\slc}^{\agg}-L_{\slc}^{\sep}\cdot \sqrt{\frac{1}{\eta^{\sep}_K}}.
\end{align*}}
For any finite concept class $\mathcal{F}\subset \{f: X\to \{0, 1\}\}$, and the sample set $S=\{\x_1, ..., \x_N\}$, the Rademacher complexity is upper bounded by $\sqrt{\frac{2d\log(N)}{N}}$ where $d$ is the VC dimension of $\mathcal{F}$. To achieve ${\pb^{\sep}}_{\slc}<{\pb^{\agg}}_{\slc}$, we simply need to find the condition of $K$ (or $\eta^{\sep}_K$) that satisfies the below in-equation:
{\begin{align*}
    {\pb^{\sep}}_{\slc}<{\pb^{\agg}}_{\slc} &\Longrightarrow \quad 2\sqrt{\frac{2N}{\log(1/\delta)}}\frac{L_{\slc}^{\sep}-L_{\slc}^{\agg}}{\overline{\ell}-\underline{\ell}}\cdot L \cdot \sqrt{\frac{2d\log(N)}{N}}<\left(L_{\slc}^{\agg}-L_{\slc}^{\sep}\cdot \sqrt{\frac{1}{\eta_K^{\sep}}}\right) \\
    &\Longrightarrow \quad 4\frac{L_{\slc}^{\sep}-L_{\slc}^{\agg}}{\overline{\ell}-\underline{\ell}}\cdot L \cdot \sqrt{\frac{d\log(N)}{\log(1/\delta)}}<\left(L_{\slc}^{\agg}-L_{\slc}^{\sep}\cdot \sqrt{\frac{1}{\eta_{K}^{\sep}}}\right) \\
    &\Longrightarrow \quad (L_{\slc}^{\sep}-L_{\slc}^{\agg}) \cdot \underbrace{\frac{4}{\overline{\ell}-\underline{\ell}}\cdot L \cdot \sqrt{\frac{d\log(N)}{\log(1/\delta)}}}_{\text{denoted as }\alpha_{\slc}, \text{which is a function of }N, \delta, d, L}<\left(L_{\slc}^{\agg}-L_{\slc}^{\sep}\cdot \sqrt{\frac{1}{\eta_K^{\sep}}}\right) \\
    &\Longrightarrow \quad (L_{\slc}^{\sep}-L_{\slc}^{\agg}) \cdot \alpha_{\slc} <\left(L_{\slc}^{\agg}-L_{\slc}^{\sep}\cdot \sqrt{\frac{1}{\eta_K^{\sep}}}\right) \\
    &\Longrightarrow \quad (L_{\slc}^{\sep}-L_{\slc}^{\agg}) \cdot \alpha_{\slc} <\left(L_{\slc}^{\agg}-L_{\slc}^{\sep}\right) + \left(1- \sqrt{\frac{1}{\eta_K^{\sep}}} \right)\cdot L_{\slc}^{\sep} \\
    &\Longrightarrow \quad \alpha_{\slc} + 1 <  \left(1- \sqrt{\frac{1}{\eta^{\sep}_{K}}} \right)\cdot \frac{L_{\slc}^{\sep}}{ L_{\slc}^{\sep}-L_{\slc}^{\agg}} \\
    &\Longrightarrow \quad \frac{1}{\gamma}<  \left(1- (\eta^{\sep}_{K})^{-\frac{1}{2}} \right)\cdot \frac{L_{\slc}^{\sep}}{L_{\slc}^{\sep}-L_{\slc}^{\agg}} \\
   &\Longrightarrow \quad \alpha_K \cdot \frac{1}{1-(\eta^{\sep}_K)^{-\frac{1}{2}}}\leq \gamma,
\end{align*}}
where we denote by $\alpha_K:= 1-{ L_{\slc}^{\agg}}/{L_{\slc}^{\sep}} $, $\gamma = 1/(1+ {\frac{4L}{\overline{\ell}-\underline{\ell}} \sqrt{\frac{d\log(N)}{\log(1/\delta)}}})$.
\end{proof}

\subsection{Proof of Theorem \ref{thm:var_lc}}
\begin{proof}
{
\begin{align*}
&\text{Var}(\hf_{\slc}^{\uni})=\E_{(X, \nY^{\uni})\sim \nD^{\uni}}\left[\ell(\hf_{\slc}^{\uni}(X),\nY^{\uni})- \E_{(X, \nY^{\uni})\sim \nD^{\uni}}[\ell(\hf_{\slc}^{\uni}(X),\nY^{\uni})]\right]^2\\
    =&\E_{\nD^{\uni}}\Bigg[\left[\ell(\hf_{\slc}^{\uni}(X),\nY^{\uni})\right]^2+\left[\E_{\nD^{\uni}}[\ell(\hf_{\slc}^{\uni}(X),\nY^{\uni})]\right]^2 -2\ell(\hf_{\slc}^{\uni}(X),\nY^{\uni})\E_{\nD^{\uni}}[\ell(\hf_{\slc}^{\uni}(X),\nY^{\uni})]\Bigg]\\
    =&\E_{\nD^{\uni}}\left[\ell(\hf_{\slc}^{\uni}(X),\nY^{\uni})\right]^2+\left[\E_{\nD^{\uni}}[\ell(\hf_{\slc}^{\uni}(X),\nY^{\uni})]\right]^2 -\E_{\nD^{\uni}}\Big[2\ell(\hf_{\slc}^{\uni}(X),\nY^{\uni})\E_{\nD^{\uni}}[\ell(\hf_{\slc}^{\uni}(X),\nY^{\uni})]\Big]\\
      =&\E_{\nD^{\uni}}\left[\ell(\hf_{\slc}^{\uni}(X),\nY^{\uni})\right]^2-\left[\E_{\nD^{\uni}}[\ell(\hf_{\slc}^{\uni}(X),\nY^{\uni})]\right]^2\\
        =&\E_{\nD^{\uni}}\left[\ell(\hf_{\slc}^{\uni}(X),\nY^{\uni})\right]^2-(R_{\ell, \nD^{\uni}}(\hf_{\slc}^{\uni}))^2.
\end{align*}}
A special case is the 0-1 loss, i.e., $\ell(\cdot)=\mathbf{1}(\cdot)$, we then have:
\begin{align*}
    \text{Var}(\hf_{\slc}^{\uni})     =&\E_{\nD^{\uni}}\left[\ell(\hf_{\slc}^{\uni}(X),\nY^{\uni})\right]^2-(R_{\ell, \nD^{\uni}}(\hf_{\slc}^{\uni}))^2\\
    =&\E_{\nD^{\uni}}\left[\ell(\hf_{\slc}^{\uni}(X),\nY^{\uni})\right]-(R_{\ell, \nD^{\uni}}(\hf_{\slc}^{\uni}))^2\\
    =&R_{\ell, \nD^{\uni}}(\hf_{\slc}^{\uni})-(R_{\ell, \nD^{\uni}}(\hf_{\slc}^{\uni}))^2,
\end{align*}
where $R_{\ell, \nD^{\uni}}(\hf_{\slc}^{\uni})\in [0, 1]$ and $g(a)=a-a^2$ is monotonically increasing when $a<\frac{1}{2}$. Note that: $R_{\ell, \nD^{\uni}}(\hf_{\slc}^{\uni})<L_{\slc0}^{\uni} \cdot (\overline{\ell}- \underline{\ell})\cdot\sqrt{\frac{\log(1/\delta)}{2\eta_K^{\uni}N}}$, when 
\begin{align*}
    & L_{\slc0}^{\uni} \cdot (\overline{\ell}- \underline{\ell})\cdot\sqrt{\frac{\log(1/\delta)}{2\eta_K^{\uni}N}}\leq \frac{1}{2}
    \Longleftrightarrow  L_{\slc0}^{\uni}(\eta_K^{\uni})^{-\frac{1}{2}}<\sqrt{\frac{N}{2(\overline{\ell}- \underline{\ell})^2\log(1/\delta)}},
\end{align*}
we have: $\text{Var}(\hf_{\slc}^{\uni}) \leq g\left(\frac{L_{\slc}^{\uni}\cdot (\overline{\ell}- \underline{\ell})}{L}\cdot\sqrt{\frac{2\log(1/\delta)}{\eta_K^{\uni}N}}\right)$. 

To achieve: $g\left(\frac{L_{\slc}^{\sep}\cdot (\overline{\ell}- \underline{\ell})}{L}\cdot\sqrt{\frac{2\log(1/\delta)}{\eta_K^{\sep}N}}\right)\leq g\left(\frac{L_{\slc}^{\agg}\cdot (\overline{\ell}- \underline{\ell})}{L}\cdot\sqrt{\frac{2\log(1/\delta)}{\eta_K^{\agg}N}}\right)$, we simply need: 
{\begin{align*}
    &L_{\slc0}^{\sep} \cdot (\overline{\ell}- \underline{\ell})\cdot\sqrt{\frac{2\log(1/\delta)}{\eta_K^{\sep}N}}\leq L_{\slc0}^{\agg} \cdot (\overline{\ell}- \underline{\ell})\cdot\sqrt{\frac{2\log(1/\delta)}{\eta_K^{\agg}N}} \Longleftrightarrow  \sqrt{\eta_K^{\uni}}>\frac{L_{\slc}^{\sep}}{L_{\slc}^{\agg}}.
\end{align*}}
\end{proof}

\subsection{Proof for Corollary~\ref{coro:multi-class_lc}}
For a general matrix $U = (T^{\uni})^{-1}$, we firstly note
\begin{align*}
        \overline{\BR_{\slc}^{\uni}}-\underline{\BR_{\slc}^{\uni}} 
    =& \max_{ i,j\in[M]} ~ U^\uni_{ij} -  \min_{ i,j\in[M]} ~  U^\uni_{ij} \\
    \le & |\max_{ i,j\in[M]} ~  U^\uni_{ij}| + | \min_{ i,j\in[M]} ~ U^\uni_{ij}| \\
    \le & |\max_{ i\in[M]} ~ \sum_{j\in[M], U_{ij}>0} U^\uni_{ij}| + | \min_{ i\in[M]} ~ \sum_{j\in[M], U_{ij}<0} ~ U^\uni_{ij}|.
\end{align*}
Recall $T^\uni \bm 1 = \bm 1 \Rightarrow \bm 1 = (T^\uni)^{-1} \bm 1$. We know the above maximum and minimum take the same $i$.
Then 
\begin{align*}
        \overline{\BR_{\slc}^{\uni}}-\underline{\BR_{\slc}^{\uni}} 
    \le & |\max_{ i\in[M]} ~ \sum_{j\in[M], U_{ij}>0} U^\uni_{ij}| + | \min_{ i\in[M]} ~ \sum_{j\in[M], U_{ij}<0} ~ U^\uni_{ij}|\\
    = & \|U^\uni\|_{\infty} \\
    \overset{(a)}{\le} & \frac{1}{\min_{i\in[M]}~ (T^{\uni}_{ii} - \sum_{j\ne i}T^{\uni}_{ij})} \\
    \le &\frac{1}{1- 2  e^{\uni}}, \quad e^\uni:=\max_{i\in[M]} (1-T_{ii}^{\uni}), \quad e^\uni < 0.5.
\end{align*}

Now we prove the inequality $(a)$ \cite{varah1975lower}.
Let $\nu$ satisfy $$\|(T^\uni)^{-1}\|_{\infty} = \|(T^\uni)^{-1} \nu\|_{\infty}/\|\nu\|_{\infty}$$ and let $\mu=(T^\uni)^{-1} \nu$.
Then 
\[
\|(T^\uni)^{-1}\|_{\infty} = \|\mu\|_{\infty}/\|\nu\|_{\infty}
\]
To bound $\|\mu\|$, we choose $i$ such that $\mu_i = \|\mu\|_{\infty}$.
Then
\[
T^\uni_{ii} \mu_i = \nu_i - \sum_{j\ne i} T^\uni_{ij} \mu_j,
\]
which further gives
\[
|T^\uni_{ii}| \|\mu\|_{\infty} \le  |\nu_i| + \sum_{j\ne i} |T^\uni_{ij}| |\mu_j| \le |\nu_i| + \|\mu\|_{\infty} \sum_{j\ne i} |T^\uni_{ij}|.
\]
Therefore,
\[
\|\mu\|_{\infty} \le \frac{|\nu_i|}{T^\uni_{ii} -  \sum_{j\ne i} T_{ij}^{\uni}},
\]
and
\[
\|(T^\uni)^{-1}\|_{\infty} = \|\mu\|_{\infty}/\|\nu\|_{\infty} \le \frac{1}{T^\uni_{ii} -  \sum_{j\ne i} T_{ij}^{\uni}}.
\]

On the other hand, denoting by $\|U\|_{\max}:=\max_{i,j\in[M]}|U_{ij}|$,  from eigenvalues, we know
\begin{align*}
    \overline{\BR_{\slc}^{\uni}}-\underline{\BR_{\slc}^{\uni}} 
    \le &  \|U^\uni\|_{\infty}\le \sqrt{M}\lambda_{\max}(U) = \frac{\sqrt{M}}{\lambda_{\min}(T^\uni)}. 
\end{align*}
where $\lambda_{\min}(T^\uni)$ denotes the minimal eigenvalue of the matrix $T^\uni$.
Therefore, 
\[
 \overline{\BR_{\slc}^{\uni}}-\underline{\BR_{\slc}^{\uni}}=L_{\slc0}^{\circ} = \min \{\frac{1}{1- 2  e_{i_{\text{max}}}^\uni},\frac{\sqrt{M}}{\lambda_{\min}(T^\uni)}\},
\]
where $e^\uni:=\max_{i\in[M]} (1-T_{ii}^{\uni}), \quad e^\uni < 0.5$, and $\lambda_{\min}(T^\uni)$ denotes the minimal eigenvalue of the matrix $T^\uni$.

\subsection{Proof of Lemma \ref{lm:distribution_shift}}
\begin{proof}
Note that for $\hf=\hf^{\uni}$, we have:
\begin{align}\label{eqn:ds}
   &R_{\ell, \D}(\hf^{\uni})-\min_{f\in \mathcal{F}}R_{\ell, \D}(f)=R_{\ell, \D}(\hf^{\uni})-R_{\ell, \D}(f^*)\notag \\
   =&\underbrace{R_{\ell, \D}(\hf^{\uni})-R_{\ell, \nD^{\uni}}(\hf^{\uni})}_{\text{Distribution shift}} +\underbrace{R_{\ell,\nD^{\uni}}(\hf^{\uni})-  \min_{f\in\mathcal F} R_{\ell, \nD^\uni}(f) + { \min_{f\in\mathcal F} R_{\ell, \nD^\uni}(f)-  R_{\ell, \D}(f^*) }.}_{\text{Estimation error}}
\end{align}
The term of distribution shift can be upper bounded by:
\begin{align*}
    &R_{\ell, \D}(\hf^{\uni})-R_{\ell,\nD^{\uni}}(\hf^{\uni})\\
    =& \E_{(X,Y)\sim \D} \left[\ell(\hf^{\uni}(X), Y)\right]  - \E_{(X,\nY_i^{\uni})\sim \nD^{\uni}}\left[\ell(\hf^{\uni}(X), \nY_i^{\uni})\right]\\
    \leq& \max_{f\in\mathcal{F}}\left|\E_{(X,Y)\sim \D} \left[\ell(f(X), Y)\right] - \E_{(X,\nY_i^{\uni})\sim \nD^{\uni}}\left[\ell(f(X), \nY_i^{\uni})\right]\right|\\
    =& \max_{f\in\mathcal{F}}\Big|\E_{(X,Y=1)\sim \D} \left[\ell(f(X), 1)\right] +\E_{(X,Y=0)\sim \D} \left[\ell(f(X), 0)\right]\\
    &- \E_{(X,\nY_i^{\uni})\sim \nD^{\uni},Y=1}\left[\ell(f(X), \nY_i^{\uni})\right]-\E_{(X,\nY_i^{\uni})\sim \nD^{\uni},Y=0}\left[\ell(f(X), \nY_i^{\uni})\right]\Big|\\
    =& \max_{f\in\mathcal{F}}\Big|\E_{(X,Y=1)\sim \D} \left[\ell(f(X), 1)\right] +\E_{(X,Y=0)\sim \D} \left[\ell(f(X), 0)\right]\\
    &- \E_{(X,\nY_i^{\uni}=1)\sim \nD^{\uni},Y=1}\left[\ell(f(X), 1)\right]-\E_{(X,\nY_i^{\uni}=0)\sim \nD^{\uni},Y=1}\left[\ell(f(X), 0)\right]\\
    &- \E_{(X,\nY_i^{\uni}=1)\sim \nD^{\uni},Y=0}\left[\ell(f(X), 1)\right]-\E_{(X,\nY_i^{\uni}=0)\sim \nD^{\uni},Y=0}\left[\ell(f(X), 0)\right]\Big|\\
    =& \max_{f\in\mathcal{F}}\Big|\underline{\E_{(X,Y=1)\sim \D} \left[\ell(f(X), 1)\right] }+\underline{\E_{(X,Y=0)\sim \D} \left[\ell(f(X), 0)\right]}\\
    &- \underline{\E_{(X,Y=1)\sim \D}\left[\P(\nY_i^{\uni}=1|Y=1)\cdot \ell(f(X), 1)\right]}-\E_{(X,Y=1)\sim \D}\left[\P(\nY_i^{\uni}=0|Y=1)\cdot \ell(f(X), 0)\right]\\
    &- \E_{(X,Y=0)\sim \D}\left[\P(\nY_i^{\uni}=1|Y=0)\cdot \ell(f(X), 1)\right]-\underline{\E_{(X,Y=0)\sim \D}\left[\P(\nY_i^{\uni}=0|Y=0)\cdot \ell(f(X), 0)\right]}\Big|.\\
    &\text{Combine similar terms, we then have:}\\
    =& \max_{f\in\mathcal{F}}\Big|\E_{(X,Y=1)\sim \D} 
    \left[\P(\nY_i^{\uni}=0|Y=1)\cdot \ell(f(X), 1)\right]+\E_{(X,Y=0)\sim \D} \left[\P(\nY_i^{\uni}=1|Y=0)\cdot \ell(f(X), 0)\right]\\
    &-\E_{(X,Y=1)\sim \D}\left[\P(\nY_i^{\uni}=0|Y=1)\cdot \ell(f(X), 0)\right]- \E_{(X,Y=0)\sim \D}\left[\P(\nY^{\uni}_i=1|Y=0)\cdot \ell(f(X), 1)\right]\Big|\\
    =& \max_{f\in\mathcal{F}}\Big|\E_{(X,Y=1)\sim \D} 
    \left[\rho^{\uni}_1\cdot \left(\ell(f(X), 1)-\ell(f(X), 0)\right)\right]+\E_{(X,Y=0)\sim \D} \left[\rho^{\uni}_0\cdot \left(\ell(f(X), 0)-\ell(f(X), 1)\right)\right]\Big|\\
    \leq& \max_{f\in\mathcal{F}}\Big|\E_{(X,Y=1)\sim \D} 
    \left[\rho^{\uni}_1\cdot \left(\overline{\ell}-\underline{\ell}\right)\right]+\E_{(X,Y=0)\sim \D} \left[\rho^{\uni}_0\cdot \left(\overline{\ell}-\underline{\ell}\right)\right]\Big|\\
    = &(p_1\rho^{\uni}_1+p_0\rho^{\uni}_0)\cdot \left(\overline{\ell}-\underline{\ell}\right).
\end{align*}
Thus, we have:
\begin{align*}
R_{\ell, \D}(\hf)-R_{\ell, \nD^{\uni}}(\hf) \le \pb^{\uni,1} := (\rho_0^{\uni} p_0+\rho_1^{\uni} p_1)\cdot\left(\overline{\ell}-\underline{\ell}\right).
\end{align*}
\end{proof}

\subsection{Proof of Lemma \ref{lm:estimation_error}}
\begin{proof}
For the term Estimation error, we have:
\begin{align*}
 &R_{\ell,\nD^{\uni}}(\hf)-  R_{\ell, \D}(f^*)\\=
 &\underbrace{R_{\ell,\nD^{\uni}}(\hf^{\uni})-  \min_{f\in\mathcal F} R_{\ell, \nD^\uni}(f)  + { \min_{f\in\mathcal F} R_{\ell, \nD^\uni}(f)-  R_{\ell, \D}(f^*) }}_{\text{Estimation error}}\\
    \leq & \underbrace{R_{\ell,\nD^{\uni}}(\hf^{\uni})-  \min_{f\in\mathcal F} R_{\ell, \nD^\uni}(f)}_{\text{Error 1}} + \underbrace{|{ \min_{f\in\mathcal F} R_{\ell, \nD^\uni}(f)-  R_{\ell, \D}(f^*) }|}_{\text{Error 2}}
\end{align*}

The upper bound of Error 1 could be derived directly from the proof of Lemma \ref{lm:bound}: since the loss function makes no use of loss correction, the L-Lipschitz constant does not have to multiply with the constant and $L_{\slc}^{\uni}\to L$. Besides, the constant for the variance term (square term) reduces to $(\overline{\ell}-\underline{\ell})$. Thus, we have:
\begin{align*}
    \text{Error 1}\leq 4L \Rk(\mathcal{F})+(\overline{\ell}- \underline{\ell})\cdot\sqrt{\frac{2\log(1/\delta)}{{\eta^{\uni}_K}N}}, ~~\forall f \in \mathcal F.
\end{align*}
For the term Error 2, the upper bound could be derived with the same procedure as adopted in the proof of Lemma \ref{lm:distribution_shift}. Thus, we obtain:
\begin{align*}
   R_{\ell,\nD^{\uni}}(\hf)-R_{\ell, \D}(f^*) \le  \underbrace{4L \Rk(\mathcal{F})+(\overline{\ell}- \underline{\ell})\cdot\sqrt{\frac{2\log(1/\delta)}{{\eta^{\uni}_K}N}}+ \pb^{\uni,1}}_{\text{Defined as: }\pb^{\uni,2}}.
\end{align*}
\end{proof}

\subsection{Proof of Theorem \ref{tm:bounds_ce}}
\begin{proof}
To achieve a smaller upper bound for the separation method, mathematically, we want:
\begin{align*}
    &4L \Rk(\mathcal{F})+(\overline{\ell}- \underline{\ell})\cdot\sqrt{\frac{2\log(1/\delta)}{{\eta^{\sep}_K}N}}+2(\rho^{\sep}_0 p_0+\rho^{\sep}_1 p_1)\cdot\left(\overline{\ell}-\underline{\ell}\right) \\
    \leq &4L \Rk(\mathcal{F})+(\overline{\ell}- \underline{\ell})\cdot\sqrt{\frac{2\log(1/\delta)}{{\eta^{\agg}_K}N}}+2(\rho^{\agg}_0 p_0+\rho^{\agg}_1 p_1)\cdot\left(\overline{\ell}-\underline{\ell}\right),
    \end{align*}
which is equivalent to prove:
\begin{align}\label{eqn:req_ce_upper}
    &\sqrt{\frac{\log(1/\delta)}{2N}} ((\eta^{\sep}_K)^{-\frac{1}{2}}-1) \cdot  (\overline{\ell}- \underline{\ell})\leq \left[(\rho^{\agg}_0 p_0+\rho^{\agg}_1 p_1)-(\rho^{\sep}_0 p_0+\rho^{\sep}_1 p_1)\right] \cdot  \left(\overline{\ell}-\underline{\ell}\right)\notag\\
    \Longleftrightarrow&\sqrt{\frac{\log(1/\delta)}{2N}} (1-(\eta^{\sep}_K)^{-\frac{1}{2}})\geq \underbrace{\left[(\rho^{\sep}_0 p_0+\rho^{\sep}_1 p_1)-(\rho^{\agg}_0 p_0+\rho^{\agg}_1 p_1)\right].}_{\text{De-noising effect of aggregation }\geq 0}
\end{align}
Eqn. (\ref{eqn:req_ce_upper}) then requires: $\sqrt{\frac{\log(1/\delta)}{2N}} \geq \frac{(\rho^{\sep}_0 p_0+\rho^{\sep}_1 p_1)-(\rho^{\agg}_0 p_0+\rho^{\agg}_1 p_1)}{(1-(\eta^{\sep}_K)^{-\frac{1}{2}})} $, which is mentioned as $\alpha_K \cdot \frac{1}{1-(\eta^{\sep}_K)^{-\frac{1}{2}}}\leq \gamma,$ where  $\alpha_K:=(\rho^{\sep}_0 p_0+\rho^{\sep}_1 p_1)-(\rho^{\agg}_0 p_0+\rho^{\agg}_1 p_1)$, $\gamma = \sqrt{{\log(1/\delta)}/{2N}}$.

\end{proof}

\subsection{Proof of Theorem \ref{thm:var_l}}
\begin{proof}
For $\uni\in\{\sep, \agg\}$, we have:
\begin{align*}
\text{Var}(\hf^{\uni})
=&\E_{(X, \nY^{\uni})\sim \nD^{\uni}}\left[\ell(\hf^{\uni}(X),\nY^{\uni})- \E_{(X, \nY^{\uni})\sim \nD^{\uni}}[\ell(\hf^{\uni}(X),\nY^{\uni})]\right]^2\\
    =&\E_{\nD^{\uni}}\Bigg[\left[\ell(\hf^{\uni}(X),\nY^{\uni})\right]^2+\left[\E_{\nD^{\uni}}[\ell(\hf^{\uni}(X),\nY^{\uni})]\right]^2-2\ell(\hf^{\uni}(X),\nY^{\uni})\E_{\nD^{\uni}}[\ell(\hf^{\uni}(X),\nY^{\uni})]\Bigg]\\
    =&\E_{\nD^{\uni}}\left[\ell(\hf^{\uni}(X),\nY^{\uni})\right]^2+\left[\E_{\nD^{\uni}}[\ell(\hf^{\uni}(X),\nY^{\uni})]\right]^2 -\E_{\nD^{\uni}}\Big[2\ell(\hf^{\uni}(X),\nY^{\uni})\E_{\nD^{\uni}}[\ell(\hf^{\uni}(X),\nY^{\uni})]\Big]\\
      =&\E_{\nD^{\uni}}\left[\ell(\hf^{\uni}(X),\nY^{\uni})\right]^2-\left[\E_{\nD^{\uni}}[\ell(\hf^{\uni}(X),\nY^{\uni})]\right]^2\\
        =&\E_{\nD^{\uni}}\left[\ell(\hf^{\uni}(X),\nY^{\uni})\right]^2-(R_{\ell, \nD^{\uni}}(\hf^{\uni}))^2.
\end{align*}
A special case is the 0-1 loss, i.e., $\ell(\cdot)=\mathbf{1}(\cdot)$, we then have:
\begin{align*}
    \text{Var}(\hf^{\uni})     =&\E_{\nD^{\uni}}\left[\ell(\hf^{\uni}(X),\nY^{\uni})\right]^2-(R_{\ell, \nD^{\uni}}(\hf^{\uni}))^2\\
    =&\E_{\nD^{\uni}}\left[\ell(\hf^{\uni}(X),\nY^{\uni})\right]-(R_{\ell, \nD^{\uni}}(\hf^{\uni}))^2\\
    =&R_{\ell, \nD^{\uni}}(\hf^{\uni})-(R_{\ell, \nD^{\uni}}(\hf^{\uni}))^2=g\left(R_{\ell, \nD^{\uni}}(\hf^{\uni})\right).
\end{align*}
where $R_{\ell, \nD^{\uni}}(\hf^{\uni})\in [0, 1]$ and $g(a)=a-a^2$ is monotonically increasing when $a<\frac{1}{2}$. Thus, when 
\begin{align*}
    R_{\ell, \nD^{\uni}}(\hf^{\uni})\leq \underbrace{(\overline{\ell}- \underline{\ell})}_{\text{reduces to 1}}\cdot\sqrt{\frac{\log(1/\delta)}{2{\eta^{\uni}_K}N}}\leq\frac{1}{2}\Longleftrightarrow \eta^{\uni}_K\geq \frac{2\log(1/\delta)}{N},
\end{align*}
we could derive $\text{Var}(\hf^{\uni})   \leq g(\sqrt{\frac{2\log(1/\delta)}{\eta^{\uni}_K N}})$.

\end{proof}

\subsection{Proof of Corollary \ref{coro:multi-class_ce}}
\begin{proof}
In the multi-class extension, the only difference is the upper bound of Distribution Shift term in Eqn. (\ref{eqn:ds}), which now becomes:
\begin{align*}
    &R_{\ell, \D}(\hf^{\uni})-R_{\ell,\nD^{\uni}}(\hf^{\uni})
    \\
     = &\E_{(X,Y)\sim \D} \left[\ell(\hf^{\uni}(X), Y)\right] - \E_{(X,\nY^{\uni})\sim \nD^{\uni}}\left[\ell(\hf^{\uni}(X), \nY^{\uni})\right]\\
    \leq& \max_{f\in\mathcal{F}}\Bigg|\E_{(X,Y)\sim \D} \left[\ell(f(X), Y)\right] - \E_{(X,\nY^{\uni})\sim \nD^{\uni}}\left[\ell(f(X), \nY^{\uni})\right]\Bigg|\\
    =& \max_{f\in\mathcal{F}}\left|\left[\sum_{j\in[M]}\E_{(X,Y=j)\sim \D} \Bigg[\ell(f(X), j)\right]\right] -\left[\sum_{j\in[M]}\E_{(X,\nY^{\uni})\sim \nD^{\uni},Y=j}\left[\ell(f(X), \nY^{\uni})\right]\right]\Bigg|\\
    =& \max_{f\in\mathcal{F}}\Bigg|\left[\sum_{j\in[M]}\E_{(X,Y=j)\sim \D} \left[\ell(f(X), j)\right]\right] -\left[ \sum_{k\in[M]}\sum_{j\in[M]}\E_{(X,Y=j)\sim \D}\left[\P(\nY^{\uni}=k|Y=j)\cdot \ell(f(X), k)\right]\right]\Bigg|\\
    =& \max_{f\in\mathcal{F}}\Bigg|\left[\sum_{j\in[M]}\E_{(X,Y=j)\sim \D} \left[\P(\nY^{\uni}\neq j|Y=j)\cdot \ell(f(X), j)\right]\right]\\
    &\qquad -\left[ \sum_{k\in[M],k\neq j}\sum_{j\in[M]}\E_{(X,Y=j)\sim \D}\left[\P(\nY^{\uni}=k|Y=j)\cdot \ell(f(X), k)\right]\right]\Bigg|\\
     =& \max_{f\in\mathcal{F}}\Bigg|\sum_{j\in[M]}\E_{(X,Y=j)\sim \D} \Bigg[\P(\nY^{\uni}\neq j|Y=j)\cdot \ell(f(X), j)\ - \sum_{k\in[M],k\neq j}\P(\nY^{\uni}=k|Y=j)\cdot \ell(f(X), k)\Bigg]\Bigg|\\
  \leq& \max_{f\in\mathcal{F}}\Bigg|\sum_{j\in[M]}\E_{(X,Y=j)\sim \D} \Bigg[\P(\nY^{\uni}\neq j|Y=j)\cdot\left( \overline{\ell}-\underline{\ell}\right)  \Bigg]\Bigg|\qquad \text{(Assumed uniform prior)} \\
  = & \sum_{j\in[M]}\P(Y=j)\cdot(1-T^{\uni}_{j,j})\left( \overline{\ell}-\underline{\ell}\right).\ \\
\end{align*}
\end{proof}

\subsection{Proof of Lemma \ref{lm:bound_pl}}
\begin{proof}
The proof of Lemma \ref{lm:bound_pl} builds on Theorem 7 in \cite{liu2020peer}: The performance bound for aggregation methods is the special case of Theorem 7 in \cite{liu2020peer} (adopting $\alpha^*=1$ defined in \cite{liu2020peer}). As for that of separation methods, the incurred difference lies in the appearance of the weight of sample complexity $\eta^{\sep}_{K}$.
Thus, we have:
\begin{align*}
   & R_{\ell, \D}(\hf^{\uni}_{\spl})-R_{\ell, \D}(f^*)
    \leq\frac{1}{1-\rho_0^{\uni}-\rho_1^{\uni}}\left( 8L \Rk(\mathcal{F})+\sqrt{\frac{2\log(4/\delta)}{ \eta^{\uni}_K N}}\left(1+2(\bar{\ell}-\underline{\ell})\right)\right)\\
    \Longleftrightarrow &R_{\ell, \D}(\hf_{\spl}^{\uni})- R_{\ell, \D}(f^*) \leq {\pb^{\uni}}_{\spl},
\end{align*}
where ${\pb^{\uni}}_{\spl}:=8L_{\spl}^{\uni} \Rk(\mathcal{F})+L_{\spl0}^{\uni}\sqrt{\frac{2\log(4/\delta)}{ \eta^{\uni}_K N}}\left(1+2(\bar{\ell}-\underline{\ell})\right)$.

\end{proof}

\subsection{Proof of Theorem \ref{tm:bounds_pl}}
\begin{proof}
Denote by ${\pb^{\uni}}_{\spl}:=\frac{8L \Rk(\mathcal{F})}{{1-\rho^{\uni}_0-\rho^{\uni}_1}}+\frac{4\sqrt{\frac{\log(4/\delta)}{2{\eta^{\uni}_{K}}N}}\left(1+2(\bar{\ell}-\underline{\ell})\right)}{{1-\rho^{\uni}_0-\rho^{\uni}_1}}$, in order to achieve ${\pb^{\sep}}_{\spl}<{\pb^{\agg}}_{\spl}$, we require $ {\pb^{\sep}}_{\spl}<{\pb^{\agg}}_{\spl}$, which is equivalent to:
{\begin{align*}
    &\frac{8L \Rk(\mathcal{F})}{{1-\rho^{\sep}_0-\rho^{\sep}_1}}+\frac{4\sqrt{\frac{\log(4/\delta)}{2{\eta^{\sep}_{K}}N}}\left(1+2(\bar{\ell}-\underline{\ell})\right)}{{1-\rho^{\sep}_0-\rho^{\sep}_1}}<\frac{8L \Rk(\mathcal{F})}{{1-\rho^{\agg}_0-\rho^{\agg}_1}} +\frac{4\sqrt{\frac{\log(4/\delta)}{2N}}\left(1+2(\bar{\ell}-\underline{\ell})\right)}{{1-\rho^{\agg}_0-\rho^{\agg}_1}},\end{align*}}
which is further equivalent to:
{\begin{align*}
    & \frac{8L \Rk(\mathcal{F})}{1-\rho^{\sep}_0-\rho^{\sep}_1}-\frac{8L \Rk(\mathcal{F})}{1-\rho^{\agg}_0-\rho^{\agg}_1}
    <  \frac{4\sqrt{\frac{\log(4/\delta)}{2N}}\left(1+2(\bar{\ell}-\underline{\ell})\right)}{1-\rho^{\agg}_0-\rho^{\agg}_1}-\frac{4\sqrt{\frac{\log(4/\delta)}{2\eta^{\sep}_{K}N}}\left(1+2(\bar{\ell}-\underline{\ell})\right)}{1-\rho^{\sep}_0-\rho^{\sep}_1}.
\end{align*}}
Note that both $1-\rho^{\sep}_0-\rho^{\sep}_1$ and $1-\rho^{\agg}_0-\rho^{\agg}_1$ are positive, the above requirement then reduces to:
{\begin{align*}
    &[(\rho^{\sep}_0+\rho^{\sep}_1)-(\rho^{\agg}_0+\rho^{\agg}_1)]8L \Rk(\mathcal{F})<\left[(1-\rho^{\sep}_0-\rho^{\sep}_1)-(1-\rho^{\agg}_0-\rho^{\agg}_1)\sqrt{\frac{1}{\eta^{\sep}_{K}}}\right]4\sqrt{\frac{\log(4/\delta)}{2N}}\left(1+2(\bar{\ell}-\underline{\ell})\right)\\
     \Longleftrightarrow &\frac{[(\rho^{\sep}_0+\rho^{\sep}_1)-(\rho^{\agg}_0+\rho^{\agg}_1)]8L \Rk(\mathcal{F})}{4\sqrt{\frac{\log(4/\delta)}{2N}}\left(1+2(\bar{\ell}-\underline{\ell})\right)} <(1-\rho^{\sep}_0-\rho^{\sep}_1)-(1-\rho^{\agg}_0-\rho^{\agg}_1)\sqrt{\frac{1}{\eta^{\sep}_{K}}}.
\end{align*}}
Note that for any finite concept class $\mathcal{F}\subset \{f: X\to \{0, 1\}\}$, and the sample set $S=\{\x_1, ..., \x_N\}$, the Rademacher complexity is upper bounded by $\sqrt{\frac{2d\log(N)}{N}}$ where $d$ is the VC dimension of $\mathcal{F}$,
a more strict condition to get becomes:
{
\begin{align*}
     \sqrt{\frac{1}{\eta^{\sep}_{K}}} <\frac{(1-\rho^{\sep}_0-\rho^{\sep}_1)}{(1-\rho^{\agg}_0-\rho^{\agg}_1)}-\frac{[(\rho^{\sep}_0+\rho^{\sep}_1)-(\rho^{\agg}_0+\rho^{\agg}_1)]8L {{\sqrt{\frac{2d\log(N)}{N}}}}}{4(1-\rho^{\agg}_0-\rho^{\agg}_1)\sqrt{\frac{\log(4/\delta)}{2N}}\left(1+2(\bar{\ell}-\underline{\ell})\right)}.
\end{align*}}
Denote by $\alpha_K:= 1-{ L_{\spl}^{\agg}}/{L_{\spl}^{\sep}} $, $\gamma =  \frac{1+2(\bar{\ell}-\underline{\ell})}{2L}\sqrt{\frac{\log(4/\delta)}{4d\log(N)} } $. The above condition is sastisfied if and only if
\begin{align*}
  \alpha_K \cdot \frac{1}{{ L_{\spl}^{\agg}}/{L_{\spl}^{\sep}}-(\eta^{\sep}_K)^{-\frac{1}{2}}}\leq \gamma. 
\end{align*}
\end{proof}

\subsection{Proof of Theorem \ref{thm:var_pl}}
\begin{proof}
Similar to the proof of Theorem \ref{thm:var_l}, for $\uni\in\{\sep, \agg\}$, we have:
\begin{align*}
&\text{Var}(\hf_{\spl}^{\uni})=\E_{\nD^{\uni}}\left[\ell(\hf_{\spl}^{\uni}(X),\nY^{\uni})\right]^2-(R_{\ell, \nD^{\uni}}(\hf_{\spl}^{\uni}))^2.
\end{align*}
A special case is the 0-1 loss, i.e., $\ell(\cdot)=\mathbf{1}(\cdot)$, we then have:
\begin{align*}
    \text{Var}(\hf_{\spl}^{\uni})     =&\E_{\nD^{\uni}}\left[\ell(\hf_{\spl}^{\uni}(X),\nY^{\uni})\right]^2-(R_{\ell, \nD^{\uni}}(\hf_{\spl}^{\uni}))^2\\
    =&\E_{\nD^{\uni}}\left[\ell(\hf_{\spl}^{\uni}(X),\nY^{\uni})\right]-(R_{\ell, \nD^{\uni}}(\hf_{\spl}^{\uni}))^2\\
    =&R_{\ell, \nD^{\uni}}(\hf_{\spl}^{\uni})-(R_{\ell, \nD^{\uni}}(\hf_{\spl}^{\uni}))^2=g\left(R_{\ell, \nD^{\uni}}(\hf_{\spl}^{\uni})\right)
\end{align*}
where $R_{\ell, \nD^{\uni}}(\hf_{\spl}^{\uni})\in [0, 1]$ and $g(a)=a-a^2$ is monotonically increasing when $a<\frac{1}{2}$. Note that: $$R_{\ell, \nD^{\uni}}(\hf_{\spl}^{\uni})<\frac{1}{1-\rho_0^{\uni}-\rho_1^{\uni}}\sqrt{\frac{\log(4/\delta)}{2 \eta^{\uni}_K N}}\left(1+2(\bar{\ell}-\underline{\ell})\right),$$ when 
\begin{align*}
    &\frac{1}{1-\rho_0^{\uni}-\rho_1^{\uni}}\sqrt{\frac{\log(4/\delta)}{2 \eta^{\uni}_K N}}\left(1+2(\bar{\ell}-\underline{\ell})\right)\leq \frac{1}{2}\\
    \Longleftrightarrow  &\sqrt{\eta_K^{\uni}}\geq  \sqrt{\frac{2\log(4/\delta)}{N}}\frac{1+2(\bar{\ell}-\underline{\ell})}{1-\rho_0^{\uni}-\rho_1^{\uni}},
\end{align*}
we have: $\text{Var}(\hf_{\slc}^{\uni}) \leq g\left(\sqrt{\frac{\log(4/\delta)}{2 \eta^{\uni}_K N}}\frac{1+2(\bar{\ell}-\underline{\ell})}{1-\rho_0^{\uni}-\rho_1^{\uni}}\right)$. To achieve: $\text{Var}(\hf^{\sep}_{\spl})<\text{Var}(\hf^{\agg}_{\spl})$, we simply need: 
{\begin{align*}
    &\sqrt{\frac{\log(4/\delta)}{2 \eta^{\sep}_K N}}\frac{1+2(\bar{\ell}-\underline{\ell})}{1-\rho_0^{\sep}-\rho_1^{\sep}}\leq \sqrt{\frac{\log(4/\delta)}{2 \eta^{\agg}_K N}}\frac{1+2(\bar{\ell}-\underline{\ell})}{1-\rho_0^{\agg}-\rho_1^{\agg}}\Longleftrightarrow \sqrt{\eta_K^{\sep}}\geq \frac{L_{\spl0}^{\sep}}{L_{\spl0}^{\agg}}.
\end{align*}}

\end{proof}

\subsection{Proof of Corollary \ref{coro:multi-class_pl}}
\begin{proof}
Regarding the multi-class extension of Lemma \ref{lm:bound_pl}, the only different thing lies in the constant: $L_{\spl0}^{\uni}$. The following Lemma \ref{lm:affine} helps us find out the multi-class form of $L_{\spl0}^{\uni}$.
\begin{lemma} \label{lm:affine}
Assume the clean label $Y$ has equal prior $P(Y=j)=\frac{1}{M}, \forall j\in [M]$. For the uniform noise transition matrix  \cite{wei2020optimizing} such that $T^{\uni}_{i,j}=\rho_i^{\uni}, \forall j\in[M]$, the expected $\ell_{\spl}$ in the multi-class setting is invariant to label noise up to an affine transformation: 
\begin{equation}\label{noisefree}
\mathbb{E}_{(X,\nY^{\uni})\sim\nD^{\uni}}[\ell_{\spl}(f(X), \nY^{\uni})] = \left(1-\sum_{j \in [M]} {\rho_j^{\uni}}\right) \mathbb{E}_\mathcal{D}[\ell_{\spl}(f(X), Y)].
\end{equation}
\end{lemma}
\paragraph{Proof of Lemma \ref{lm:affine}}
Recall that $\mathcal{D}$ and $\nD^{\uni}$ refer to the joint distribution over $(X, {Y})$ and $(X, \nY^{\uni})$, respectively. We further denote the marginal distributions of $X$, $Y$, and $\nY^{\uni}$ by $\mathcal D_X$, $\mathcal D_Y$, and $\nD_{\nY^{\uni}}$, respectively.
Let $X_{p}\sim \mathcal D_X$, $\nY^{\uni}_{p}\sim\nD_{\nY^{\uni}}$ be the random variables corresponding to the peer samples. The peer loss function is defined as
\begin{equation}\label{Eq:peerlossSup}
\ell_{\spl}(f(x_n), \ny^{\uni}_n) = \ell(f(x_n), \ny^{\uni}_n) - \ell(f(x_{p,n}), \tilde y^{\uni}_{p,n}),
\end{equation}
where $(x_n,\ny^{\uni}_n)$ is a normal training sample pair, $x_{p,n}$ and $\tilde y_{p,n}^{\uni}$ are corresponding peer samples.

Taking expectation for (\ref{Eq:peerlossSup}) yields
\begin{equation}\label{Eq:peerlossExpSup}
\mathbb E_{\nD^{\uni}} [\ell_{\spl}(f(X), \nY^{\uni})] = \mathbb E_{\nD^{\uni}}[\ell(f(X), \nY^{\uni})] - \mathbb E_{\widetilde{\mathcal{D}}_{\nY^{\uni}}} \left[ \mathbb E_{\mathcal{D}_X}[\ell(f(X_{p}), \nY^{\uni}_{p})]\right].
\end{equation}
The first term in (\ref{Eq:peerlossExpSup}) is
{\begin{align*}
  & \mathbb{E}_{\nD^{\uni}}[\ell(f(X), \nY^{\uni})] \\
= & \sum_{j \in [M]} \sum_{i\in[M]} T_{ij}^{\uni}\cdot  \mathbb P(Y=i) \cdot \mathbb{E}_{\mathcal D |{Y}=i}[\ell(f(X), j)]  \\
= & \sum_{j \in [M]} \Bigg[ T_{jj}^{\uni} \cdot\mathbb P(Y=j) \cdot \mathbb{E}_{\mathcal D |{Y}=j}[\ell(f(X), j)]  + 
\sum_{i\in[M], i\ne j} T_{ij}^{\uni} \cdot\mathbb P(Y=i) \cdot \mathbb{E}_{\mathcal D |{Y}=i}[\ell(f(X), j)] \Bigg] \\
= & \sum_{j \in [M]} \Bigg[ \left(1-\sum_{i\ne j,i\in[M]} T_{ji}^{\uni}\right) \cdot \mathbb P(Y=j) \cdot \mathbb{E}_{\mathcal D |{Y}=j}[\ell(f(X), j)] + 
\sum_{i\in[M], i\ne j} T_{ij}^{\uni} \cdot\mathbb P(Y=i) \cdot \mathbb{E}_{\mathcal D |{Y}=i}[\ell(f(X), j)] \Bigg].
\end{align*}}
Accordingly, noting $X_{p}$ and $\nY^{\uni}_{p}$ are independent, the second term in (\ref{Eq:peerlossExpSup}) is
{\begin{align*}
  &  \mathbb E_{\widetilde{\mathcal{D}}_{\nY^{\uni}}} \left[ \mathbb E_{\mathcal{D}_X}[\ell(f(X_{p}), \nY^{\uni}_{p})]\right] = \sum_{j \in [M]} \mathbb P(\nY^{\uni}_{p}=j) \cdot \mathbb E_{\mathcal{D}_X}[\ell(f(X_{p}), j)] \\
= & \sum_{j \in [M]} \sum_{i\in[M]} T_{ij}^{\uni}\cdot \mathbb P({Y_{p}}=i)  \cdot \mathbb{E}_{\mathcal{D}_X}[\ell(f(X), j)]  \\
= & \sum_{j \in [M]} \Bigg[ T_{jj}^{\uni} \cdot\mathbb P({Y_{p}}=j)\cdot  \mathbb{E}_{\mathcal{D}_X}[\ell(f(X), j)] + 
\sum_{i\in[M], i\ne j} T_{ij}^{\uni}\cdot \mathbb P(Y_{p}=i) \cdot \mathbb{E}_{\mathcal{D}_X}[\ell(f(X), j)] \Bigg] \\
= & \sum_{j \in [M]} \left[ \Bigg(1-\sum_{i\ne j,i\in[M]} T_{ji}^{\uni}\right) \cdot\mathbb P(Y_{p}=j) \cdot \mathbb{E}_{\mathcal{D}_X}[\ell(f(X), j)] + 
\sum_{i\in[M], i\ne j} T_{ij}^{\uni} \cdot\mathbb P(Y_{p}=i) \cdot \mathbb{E}_{\mathcal{D}_X}[\ell(f(X), j)] \Bigg].
\end{align*}}

In this case, we have $\rho^{\uni}_i = T_{ji}^{\uni}, \forall j \in [M], j\ne i$.
The first term becomes
{ \begin{align*}
  &\mathbb{E}_{\nD^{\uni}}[\ell(f(X), \nY^{\uni})] \\
  =& \sum_{j \in [M]} \Bigg[ \left(1-\sum_{i\ne j,i\in[M]} \rho^{\uni}_i\right) \cdot\mathbb P(Y=j) \cdot  \mathbb{E}_{\mathcal D |{Y}=j}[\ell(f(X), j)]  + 
\sum_{i\in[M], i\ne j} \rho_j^{\uni}\cdot \mathbb P(Y=i)  \cdot\mathbb{E}_{\mathcal D |{Y}=i}[\ell(f(X), j)] \Bigg] \\
= & \sum_{j \in [M]} \Bigg[ \left(1-\sum_{i\in[M]} \rho^{\uni}_i\right) \cdot\mathbb P(Y=j)\cdot  \mathbb{E}_{\mathcal D |{Y}=j}[\ell(f(X), j)] +
\sum_{i\in[M]} \rho_j^{\uni} \cdot\mathbb P(Y=i)\cdot  \mathbb{E}_{\mathcal D |{Y}=i}[\ell(f(X), j)] \Bigg] \\
= &  \left[ \left(1-\sum_{i\in[M]} \rho^{\uni}_i\right) \cdot\mathbb{E}_{\mathcal D}[\ell(f(X), Y)] + \sum_{j \in [M]}
 \rho_j^{\uni}\cdot  \mathbb{E}_{{\mathcal D}_X}[\ell(f(X), j)] \right].
\end{align*}}
The second term becomes
{\begin{align*}
  &  \mathbb E_{\widetilde{\mathcal{D}}_{\nY^{\uni}}} \left[ \mathbb E_{\mathcal{D}_X}[\ell(f(X_{p}), \nY^{\uni}_{p})]\right] \\
= & \sum_{j \in [M]} \Bigg[ \left(1-\sum_{i\ne j,i\in[M]} \rho^{\uni}_i\right)\cdot \mathbb P(Y_{p}=j) \cdot \mathbb{E}_{\mathcal{D}_X}[\ell(f(X), j)]  + 
\sum_{i\in[M], i\ne j} \rho^{\uni}_{j} \cdot\mathbb P(Y_{p}=i)\cdot  \mathbb{E}_{\mathcal{D}_X}[\ell(f(X), j)] \Bigg] \\
= & \sum_{j \in [M]} \Bigg[ \left(1-\sum_{i\in[M]} \rho^{\uni}_i\right)\cdot \mathbb P(Y_{p}=j) \cdot \mathbb{E}_{\mathcal{D}_X}[\ell(f(X), j)] + 
\sum_{i\in[M]} \rho^{\uni}_{j} \cdot\mathbb P(Y_{p}=i) \cdot \mathbb{E}_{\mathcal{D}_X}[\ell(f(X), j)] \Bigg] \\
= &  \left(1-\sum_{i\in[M]} \rho^{\uni}_i\right)\cdot \mathbb E_{{\mathcal{D}}_Y} \left[ \mathbb{E}_{\mathcal{D}_X}[\ell(f(X_{p}), Y_{p})] \right]+
\sum_{j \in [M]} \rho^{\uni}_{j} \cdot  \mathbb{E}_{\mathcal{D}_X}[\ell(f(X), j)].
\end{align*}}
Comparing the above two terms we have:
\begin{equation}\label{Eq:peerlossExpSup_final}
\mathbb E_{\nD^{\uni}} [\ell_{\spl}(f(X), \nY^{\uni})] = \left(1-\sum_{i\in[M]} \rho^{\uni}_i\right) \mathbb E_{ \mathcal{D}} [\ell_{\spl}(f(X), {Y})].
\end{equation}

Thus, substituting $L_{\spl0}^{\uni}:=\frac{1}{1-\rho^{\uni}_0-\rho^{\uni}_1}$ by $\frac{1}{1-\sum_{i\in[M]}\rho^{\uni}_i}$, the proof of Corollary \ref{coro:multi-class_pl} is finished if we repeat the corresponding proof of the binary task.

\end{proof}

\section{Additional Results and Details}

\section{Experiment Details}

\subsection{Experiment Details on UCI Datasets}

\paragraph{Datasets} In this paper, we conducted the experiments on two binary (Breast and German) and two multiclass (StatLog and Optical) UCI classification datasets. As for the splitting of training and testing, the original settings are used when training and testing files are provided. The remaining datasets only give one data file. We adopt 50/50 splitting for the testing results' statistical significance as more data is distributed to testing dataset. More specifically, the numbers of (training, testing) samples in Breast, German, StatLog, and Optical datasets are (285, 284), (500, 500), (4435, 2000), and (3823, 1797).

\paragraph{Generating the noisy labels on UCI datasets}

For each UCI dataset adopted in this paper, the label of each sample in the training dataset will be flipped to the other classes with the probability $\epsilon$ (noise rate). For the multiclass classification datasets, the specific label which will be flipped to is randomly selected with the equal probabilities. For binary and multiclass classification datasets, (0.1, 0.2, 0.3, 0.4) and (0.2, 0.4, 0.6, 0.8) are used as different lists of noise rates respectively. 

\paragraph{Implementation details} 
We implemented a simple two-layer ReLU Multi-Layer Perceptron (MLP) for the classification task on these four UCI datasets. 
The Adam optimizer is used with a learning rate of 0.001 and the batch size is 128.

\subsection{Experiment Details on CIFAR-10 Datasets}
The generation of symmetric noisy dataset is adopted from \cite{wei2020optimizing}. As for the instance-dependent label noise, the generating algorithm follows the state-of-the-art method \cite{xia2020part}. Both cases adopt noise rates: $[0.2, 0.4, 0.6, 0.8]$. The basic hyper-parameters settings for all methods are listed as follows: mini-batch size (128), optimizer (SGD), initial learning rate (0.1), momentum (0.9), weight decay (0.0005), number of epochs (120) and learning rate decay (0.1 at 50 epochs). Standard data augmentation is applied to each dataset. All experiments run on 8 Nvidia RTX A5000 GPUs.

\subsection{Details Results on CIFAR-10 Dataset}
Table \ref{Tab:c10_ce} includes all the detailed accuracy values appeared in Figure \ref{fig:c10}. The results on synthetic noisy CIFAR-10 dataset aligns well with the theoretical observations: label separation is preferred over label aggregation when the noise rates are high, or the number of labelers/annotations is insufficient. 
\begin{table*}[!ht]
\centering{\scalebox{0.78}{
\begin{tabular}{ccccccc||ccccccc}
\hline
\multicolumn{7}{c}{\emph{CIFAR-10, Symmetric} CE}& \multicolumn{7}{c}{\emph{CIFAR-10, Instance} CE}  \\
\hline
 $\epsilon=0.2$&  $K=3$ & $K=5$ & $K=9$ & $K=15$ & $K=25$ & $K=49$ &$\epsilon=0.2$&  $K=3$ & $K=5$ & $K=9$ & $K=15$ & $K=25$ & $K=49$\\ \hline\hline
MV  & 92.21 & \bad{92.98} & \bad{93.54} & 93.43 & \bad{93.73} & \bad{93.40} & MV   & 91.99 & \bad{93.29} & \bad{93.57} & 93.47 & \bad{93.68} & \bad{93.60}\\
EM  & 92.08 & 92.93 & \bad{93.54} & \bad{93.64} & 93.35 & 93.37 &
EM & 91.92 & 93.21 & 93.55 & \bad{93.61} & 93.44 & 93.44 \\
Sep  & \good{92.52} & 92.89 & 93.35 & 93.15 & 93.42 & \good{93.40}
& Sep & \good{92.36} & 92.97 & 93.43 & 93.24 & 93.33 & 93.35\\
 \hline
 \hline
 $\epsilon=0.4$&  $K=3$ & $K=5$ & $K=9$ & $K=15$ & $K=25$ & $K=49$ & $\epsilon=0.4$&  $K=3$ & $K=5$ & $K=9$ & $K=15$ & $K=25$ & $K=49$ \\ \hline\hline
 MV  & 89.09 & 91.59 & \bad{93.18} & 93.43 & 93.26 & 93.44& MV& 87.14 & 91.15 & 93.10 & 93.15 & 93.23 & 93.48 \\
EM  & 88.83 & 91.02 & 92.54 & \bad{93.45} & \bad{93.69} & \bad{93.68} & EM  & 88.07 & \bad{92.40} & \bad{93.70} & \bad{93.58} & \bad{93.74} & \bad{93.53} \\
Sep  & \good{90.61} & \good{91.95} & 92.70 & 92.92 & 93.32 & 93.13&
Sep& \good{90.83} & 91.90 & 92.63 & 92.46 & 93.08 & 93.26\\
 \hline
 \hline
 $\epsilon=0.6$&  $K=3$ & $K=5$ & $K=9$ & $K=15$ & $K=25$ & $K=49$ & $\epsilon=0.6$&  $K=3$ & $K=5$ & $K=9$ & $K=15$ & $K=25$ & $K=49$ \\ \hline\hline
MV  & 81.85 & 87.33 & 89.88 & \bad{91.88} & \bad{92.96} & \bad{93.40} & MV & 49.22 & 83.95 & 89.45 & 91.60 & 92.88 & \bad{93.65}\\
EM  & 81.04 & 85.91 & 89.76 & 91.57 & 92.55 & 93.10 & EM  & 78.34 & \bad{88.79} & \bad{91.95} & \bad{92.97} & \bad{93.46} & \bad{93.65}\\
Sep  & \good{87.00} & \good{89.19} & \good{90.70} & 91.97 & 92.40 & 93.17 & Sep  & \good{83.79} & 87.55 & 90.15 & 91.58 & 91.86 & 92.74\\
 \hline
 \hline
 $\epsilon=0.8$&  $K=3$ & $K=5$ & $K=9$ & $K=15$ & $K=25$ & $K=49$ & $\epsilon=0.8$&  $K=3$ & $K=5$ & $K=9$ & $K=15$ & $K=25$ & $K=49$ \\ \hline\hline
 MV  & 20.94 & 44.62 & 70.91 & 79.61 & 84.83 & 89.09 & MV  & 14.59 & 25.25 & 34.47 & 57.99 & 57.51 & 87.08 \\
EM  & 37.91 & 50.78 & 67.19 & 75.26 & 82.97 & 87.97 & EM  & 20.03 & 26.54 & \bad{65.16} & \bad{80.10} & \bad{88.59} & \bad{92.14} \\
Sep  & \good{61.47} & \good{70.10} & \good{79.61} & \good{83.93} & \good{86.82} & \good{90.04} & Sep  & \good{26.16} & \good{28.89} & 50.35 & 74.15 & 71.39 & 87.54\\
 \hline\hline
\multicolumn{7}{c}{\emph{CIFAR-10, Symmetric} BW}& \multicolumn{7}{c}{\emph{CIFAR-10, Instance} BW}  \\
\hline
 $\epsilon=0.2$&  $K=3$ & $K=5$ & $K=9$ & $K=15$ & $K=25$ & $K=49$ &$\epsilon=0.2$&  $K=3$ & $K=5$ & $K=9$ & $K=15$ & $K=25$ & $K=49$\\ \hline\hline
 MV & 92.08 & \bad{94.09} & \bad{94.92} & 94.90 & 94.79 & \bad{94.90}  & MV & \bad{92.03} & 93.87 & \bad{95.12} & \bad{95.11} & 94.97 & \bad{94.75}   \\
EM & \bad{92.13} & 93.08 & 94.90 & \bad{94.91} & \bad{94.90} & 94.86   & EM & 91.93 & \bad{94.39} & 94.90 & 94.84 & \bad{95.05} & 94.54  \\
Sep& 91.74 & 92.61 & 92.75 & 92.59 & 94.44 & 92.97 & Sep & 91.93 & 92.07 & 92.70 & 91.75 & 93.02 & 92.47  \\
 \hline
 \hline
 $\epsilon=0.4$&  $K=3$ & $K=5$ & $K=9$ & $K=15$ & $K=25$ & $K=49$ & $\epsilon=0.4$&  $K=3$ & $K=5$ & $K=9$ & $K=15$ & $K=25$ & $K=49$ \\ \hline\hline
 MV & 88.28 & \bad{91.11} & 92.73 & 94.60 & 94.62 & 94.81 & MV   & 86.61 & 90.64 & 93.00 & 94.73 & 94.72 & 94.72\\
EM  & 87.41 & 90.23 & \bad{92.83} & \bad{94.77} & \bad{94.80} & \bad{95.18}  & EM   & \bad{89.83} & \bad{92.04} & \bad{94.74} & \bad{95.00} & \bad{94.94} & \bad{94.80}\\
Sep & \good{89.14} & 89.68 & 91.07 & 92.46 & 92.26 & 94.24 & Sep  & 88.86 & 87.89 & 92.09  & 89.92 & 91.05 & 91.96  \\
 \hline
 \hline
 $\epsilon=0.6$&  $K=3$ & $K=5$ & $K=9$ & $K=15$ & $K=25$ & $K=49$ & $\epsilon=0.6$&  $K=3$ & $K=5$ & $K=9$ & $K=15$ & $K=25$ & $K=49$ \\ \hline\hline
 MV & 81.21 & 86.29 & \bad{89.51} & \bad{91.33} & \bad{93.52} & \bad{94.81} & MV  & 43.78 & 82.59 & 88.56 & 91.47 & 93.27 & \bad{95.06}  \\
EM & 78.13 & 84.33 & 89.44 & 91.17 & 92.45 & 94.60   & EM & 44.92  & \bad{}87.33 & \bad{}91.39 & \bad{}93.58 & \bad{}94.72 & 94.99   \\
Sep & \good{83.84} & \good{87.05} & 88.10 & 89.80 & 90.95 & 92.11 & Sep & \good{80.88} & 86.22 & 88.45 & 90.69 & 91.16 & 92.61 \\
 \hline
 \hline
 $\epsilon=0.8$&  $K=3$ & $K=5$ & $K=9$ & $K=15$ & $K=25$ & $K=49$ & $\epsilon=0.8$&  $K=3$ & $K=5$ & $K=9$ & $K=15$ & $K=25$ & $K=49$ \\ \hline\hline
 MV& 16.43 & 60.97 & 71.11 & \bad{}77.86 & \bad{}82.72 & \bad{}88.41  & MV  & 16.00 & 25.03 & 33.80 & 67.91 & 68.52 & 86.49 \\
EM  & 10.00  & 45.97 & 66.02 & 74.37 & 80.08 & 87.42 & EM  & 16.06 & 22.73 & \bad{}53.96 & \bad{}76.24 & \bad{}86.74 & \bad{}92.02 \\
Sep  & \good{58.48} & \good{69.86} & \good{76.03} & \good{}79.79 & 82.60 & 86.31 & Sep & \good{27.84} & \good{26.68} & 32.72 & 37.27 & 54.41 & 83.37 \\
 \hline\hline
\multicolumn{7}{c}{\emph{CIFAR-10, Symmetric} PeerLoss}& \multicolumn{7}{c}{\emph{CIFAR-10, Instance} PeerLoss}  \\
\hline
 $\epsilon=0.2$&  $K=3$ & $K=5$ & $K=9$ & $K=15$ & $K=25$ & $K=49$ &$\epsilon=0.2$&  $K=3$ & $K=5$ & $K=9$ & $K=15$ & $K=25$ & $K=49$\\ \hline\hline
 MV & 92.69 & 93.35 & \bad{}93.90 & \bad{}94.12 & \bad{}94.15 & \bad{}93.81 & MV  & 92.13 & \bad{}93.53 &\bad{} 94.00 & 93.78 & \bad{}94.13 & \bad{}94.08  \\
EM   & 92.39 & 93.25 & 93.76 & 93.93 & 93.52 & 93.77 & EM  & 91.93 & 93.51 & 93.78 & \bad{}93.88 & 94.03 & 93.82   \\
Sep & \good{}93.15 &\good{} 93.51 & 93.77 & 93.51 & 93.56 & 93.73 &  Sep  & \good{}92.86 & 93.23 & 93.56 & 93.72 & 93.63 & 93.95   \\
 \hline
 \hline
 $\epsilon=0.4$&  $K=3$ & $K=5$ & $K=9$ & $K=15$ & $K=25$ & $K=49$ & $\epsilon=0.4$&  $K=3$ & $K=5$ & $K=9$ & $K=15$ & $K=25$ & $K=49$ \\ \hline\hline
 MV & 89.40 & 91.88 & \bad{}93.42 & \bad{}93.84 & 93.83 & 94.04& MV   & 88.15 & 91.61 & 93.21 & 93.64 & 93.84 & \bad{}93.69 \\
EM  & 89.23 & 91.41 & 93.06 & \bad{}93.83 & \bad{}93.85 & \bad{}94.11  & EM  & 90.59 & 92.60 &\bad{} 93.95 & \bad{}94.02 & \bad{}94.06 & 93.68  \\
Sep & \good{}91.08 & \good{}92.38 & 93.17 & 93.40 & 93.56 & 93.37 &  Sep   &  \good{}91.06 & \good{} 92.70 & 93.22 & 92.92 & 93.65 & 93.67   \\
 \hline
 \hline
 $\epsilon=0.6$&  $K=3$ & $K=5$ & $K=9$ & $K=15$ & $K=25$ & $K=49$ & $\epsilon=0.6$&  $K=3$ & $K=5$ & $K=9$ & $K=15$ & $K=25$ & $K=49$ \\ \hline\hline
 MV& 82.88 & 87.95 & 90.42 & 92.31 & \bad{}93.61 & \bad{}93.79 & MV  & 60.66 & 84.99 & 90.30 & 91.93 & 93.16 & 93.81  \\
EM & 81.64 & 86.45 & 90.09 & 91.98 & 93.23 & 93.58   & EM  & 78.53 & \bad{}89.11 & \bad{}92.44 & \bad{}93.17 & \bad{}93.96 & \bad{}93.85  \\
Sep & \good{}87.28 &\good{} 89.80 & \good{}91.19 & \good{}92.42 & 93.18 & 93.65 &  Sep   & \good{}85.76 & 89.07 & 91.05 & 92.22 & 92.45 & 93.39   \\
 \hline
 \hline
 $\epsilon=0.8$&  $K=3$ & $K=5$ & $K=9$ & $K=15$ & $K=25$ & $K=49$ & $\epsilon=0.8$&  $K=3$ & $K=5$ & $K=9$ & $K=15$ & $K=25$ & $K=49$ \\ \hline\hline
 MV & 21.82 & 48.71 & 72.81 & 80.32 & 85.27 & 89.38 & MV & 14.35 & 24.83 & 40.49 & 65.47 & 69.28 & 88.05  \\
EM  & 38.29 & 52.63 & 68.70 & 77.42 & 83.94 & 88.45  & EM  & 26.52 & 28.43 & \bad{}66.72 & \bad{}80.71 & \bad{}89.40 & \bad{}92.41   \\
Sep & \good{}64.32 & \good{}72.52 & \good{}80.31 & \good{}84.65 &\good{} 87.40 & \good{}90.56 &  Sep    & \good{}33.87 & \good{}37.49 & 57.36 & 77.43 & 80.51 & 89.15  \\
 \hline
\end{tabular}}}
\caption{The performances of CE/BW/PeerLoss trained on (Left half: symmetric noise; right half: instance noise) CIFAR-10 aggregated labels (majority vote, EM inference), and separated labels. (Different number of labels per training image) 
}
\label{Tab:c10_ce}
\end{table*}

%% file: main.bbl
\begin{thebibliography}{10}

\bibitem{albarqouni2016aggnet}
Shadi Albarqouni, Christoph Baur, Felix Achilles, Vasileios Belagiannis,
  Stefanie Demirci, and Nassir Navab.
\newblock Aggnet: deep learning from crowds for mitosis detection in breast
  cancer histology images.
\newblock {\em IEEE transactions on medical imaging}, 35(5):1313--1321, 2016.

\bibitem{amid2019robust}
Ehsan Amid, Manfred~K Warmuth, Rohan Anil, and Tomer Koren.
\newblock Robust bi-tempered logistic loss based on {B}regman divergences.
\newblock {\em Advances in Neural Information Processing Systems}, 32, 2019.

\bibitem{bar2021multiplicative}
Noga Bar, Tomer Koren, and Raja Giryes.
\newblock Multiplicative reweighting for robust neural network optimization.
\newblock {\em arXiv preprint arXiv:2102.12192}, 2021.

\bibitem{chang2017active}
Haw-Shiuan Chang, Erik Learned-Miller, and Andrew McCallum.
\newblock Active bias: Training more accurate neural networks by emphasizing
  high variance samples.
\newblock {\em Advances in Neural Information Processing Systems}, 30, 2017.

\bibitem{chen2020structured}
Zhijun Chen, Huimin Wang, Hailong Sun, Pengpeng Chen, Tao Han, Xudong Liu, and
  Jie Yang.
\newblock Structured probabilistic end-to-end learning from crowds.
\newblock In {\em IJCAI}, pages 1512--1518, 2020.

\bibitem{cheng2020learning}
Hao Cheng, Zhaowei Zhu, Xingyu Li, Yifei Gong, Xing Sun, and Yang Liu.
\newblock Learning with instance-dependent label noise: A sample sieve
  approach.
\newblock In {\em International Conference on Learning Representations}, 2021.

\bibitem{cheng2021demystifying}
Hao Cheng, Zhaowei Zhu, Xing Sun, and Yang Liu.
\newblock Demystifying how self-supervised features improve training from noisy
  labels.
\newblock {\em arXiv preprint arXiv:2110.09022}, 2021.

\bibitem{dawid1979maximum}
Alexander~Philip Dawid and Allan~M Skene.
\newblock Maximum likelihood estimation of observer error-rates using the em
  algorithm.
\newblock {\em Journal of the Royal Statistical Society: Series C (Applied
  Statistics)}, 28(1):20--28, 1979.

\bibitem{Dua:2019}
Dheeru Dua and Casey Graff.
\newblock {UCI} machine learning repository, 2017.

\bibitem{estelles2012towards}
Enrique Estell{\'e}s-Arolas and Fernando Gonz{\'a}lez-Ladr{\'o}n-de Guevara.
\newblock Towards an integrated crowdsourcing definition.
\newblock {\em Journal of Information science}, 38(2):189--200, 2012.

\bibitem{feldman2020does}
Vitaly Feldman.
\newblock Does learning require memorization? a short tale about a long tail.
\newblock In {\em Proceedings of the 52nd Annual ACM SIGACT Symposium on Theory
  of Computing}, pages 954--959, 2020.

\bibitem{guan2018said}
Melody Guan, Varun Gulshan, Andrew Dai, and Geoffrey Hinton.
\newblock Who said what: Modeling individual labelers improves classification.
\newblock In {\em Proceedings of the AAAI Conference on Artificial
  Intelligence}, volume~32, 2018.

\bibitem{han2020survey}
Bo~Han, Quanming Yao, Tongliang Liu, Gang Niu, Ivor~W Tsang, James~T Kwok, and
  Masashi Sugiyama.
\newblock A survey of label-noise representation learning: Past, present and
  future.
\newblock {\em arXiv preprint arXiv:2011.04406}, 2020.

\bibitem{han2018co}
Bo~Han, Quanming Yao, Xingrui Yu, Gang Niu, Miao Xu, Weihua Hu, Ivor Tsang, and
  Masashi Sugiyama.
\newblock Co-teaching: Robust training of deep neural networks with extremely
  noisy labels.
\newblock In {\em Advances in neural information processing systems}, pages
  8527--8537, 2018.

\bibitem{he2016deep}
Kaiming He, Xiangyu Zhang, Shaoqing Ren, and Jian Sun.
\newblock Deep residual learning for image recognition.
\newblock In {\em Proceedings of the IEEE conference on computer vision and
  pattern recognition}, pages 770--778, 2016.

\bibitem{howe2006rise}
Jeff Howe et~al.
\newblock The rise of crowdsourcing.
\newblock {\em Wired magazine}, 14(6):1--4, 2006.

\bibitem{ipeirotis2014repeated}
Panagiotis~G Ipeirotis, Foster Provost, Victor~S Sheng, and Jing Wang.
\newblock Repeated labeling using multiple noisy labelers.
\newblock {\em Data Mining and Knowledge Discovery}, 28(2):402--441, 2014.

\bibitem{krizhevsky2009learning}
Alex Krizhevsky, Geoffrey Hinton, et~al.
\newblock Learning multiple layers of features from tiny images.
\newblock Technical report, Citeseer, 2009.

\bibitem{kumar2021constrained}
Abhishek Kumar and Ehsan Amid.
\newblock Constrained instance and class reweighting for robust learning under
  label noise.
\newblock {\em arXiv preprint arXiv:2111.05428}, 2021.

\bibitem{lecue2010sharper}
Guillaume Lecu{\'e} and Shahar Mendelson.
\newblock Sharper lower bounds on the performance of the empirical risk
  minimization algorithm.
\newblock {\em Bernoulli}, pages 605--613, 2010.

\bibitem{liu2021adaptive}
Sheng Liu, Kangning Liu, Weicheng Zhu, Yiqiu Shen, and Carlos Fernandez-Granda.
\newblock Adaptive early-learning correction for segmentation from noisy
  annotations.
\newblock {\em arXiv preprint arXiv:2110.03740}, 2021.

\bibitem{liu2020early}
Sheng Liu, Jonathan Niles-Weed, Narges Razavian, and Carlos Fernandez-Granda.
\newblock Early-learning regularization prevents memorization of noisy labels.
\newblock {\em Advances in neural information processing systems},
  33:20331--20342, 2020.

\bibitem{liu2022robust}
Sheng Liu, Zhihui Zhu, Qing Qu, and Chong You.
\newblock Robust training under label noise by over-parameterization.
\newblock {\em arXiv preprint arXiv:2202.14026}, 2022.

\bibitem{liu2016classification}
Tongliang Liu and Dacheng Tao.
\newblock Classification with noisy labels by importance reweighting.
\newblock {\em IEEE Transactions on pattern analysis and machine intelligence},
  38(3):447--461, 2016.

\bibitem{liu2021understanding}
Yang Liu.
\newblock Understanding instance-level label noise: Disparate impacts and
  treatments.
\newblock In {\em International Conference on Machine Learning}, pages
  6725--6735. PMLR, 2021.

\bibitem{liu2020peer}
Yang Liu and Hongyi Guo.
\newblock Peer loss functions: Learning from noisy labels without knowing noise
  rates.
\newblock In {\em International Conference on Machine Learning}, pages
  6226--6236. PMLR, 2020.

\bibitem{liu2015online}
Yang Liu and Mingyan Liu.
\newblock An online learning approach to improving the quality of
  crowd-sourcing.
\newblock {\em ACM SIGMETRICS Performance Evaluation Review}, 43(1):217--230,
  2015.

\bibitem{lukasik2020does}
Michal Lukasik, Srinadh Bhojanapalli, Aditya Menon, and Sanjiv Kumar.
\newblock Does label smoothing mitigate label noise?
\newblock In {\em International Conference on Machine Learning}, pages
  6448--6458. PMLR, 2020.

\bibitem{luo2020research}
Tianyi Luo, Xingyu Li, Hainan Wang, and Yang Liu.
\newblock Research replication prediction using weakly supervised learning.
\newblock In {\em In Proceedings of the 2020 Conference on Empirical Methods in
  Natural Language Processing: Findings}, 2020.

\bibitem{luo2019machine}
Tianyi Luo and Yang Liu.
\newblock Machine truth serum.
\newblock {\em arXiv preprint arXiv:1909.13004}, 2019.

\bibitem{ma2020normalized}
Xingjun Ma, Hanxun Huang, Yisen Wang, Simone Romano, Sarah Erfani, and James
  Bailey.
\newblock Normalized loss functions for deep learning with noisy labels.
\newblock In {\em International Conference on Machine Learning}, pages
  6543--6553. PMLR, 2020.

\bibitem{majidi2021exponentiated}
Negin Majidi, Ehsan Amid, Hossein Talebi, and Manfred~K. Warmuth.
\newblock Exponentiated gradient reweighting for robust training under label
  noise and beyond.
\newblock {\em arXiv preprint arXiv:2104.01493}, 2021.

\bibitem{mendelson2008lower}
Shahar Mendelson.
\newblock Lower bounds for the empirical minimization algorithm.
\newblock {\em IEEE Transactions on Information Theory}, 54(8):3797--3803,
  2008.

\bibitem{mitra2015credbank}
Tanushree Mitra and Eric Gilbert.
\newblock Credbank: A large-scale social media corpus with associated
  credibility annotations.
\newblock In {\em Proceedings of the International AAAI Conference on Web and
  Social Media}, volume~9, pages 258--267, 2015.

\bibitem{natarajan2013learning}
Nagarajan Natarajan, Inderjit~S Dhillon, Pradeep~K Ravikumar, and Ambuj Tewari.
\newblock Learning with noisy labels.
\newblock In {\em Advances in neural information processing systems}, pages
  1196--1204, 2013.

\bibitem{patrini2017making}
Giorgio Patrini, Alessandro Rozza, Aditya Krishna~Menon, Richard Nock, and
  Lizhen Qu.
\newblock Making deep neural networks robust to label noise: A loss correction
  approach.
\newblock In {\em Proceedings of the IEEE Conference on Computer Vision and
  Pattern Recognition}, pages 1944--1952, 2017.

\bibitem{pennycook2019fighting}
Gordon Pennycook and David~G Rand.
\newblock Fighting misinformation on social media using crowdsourced judgments
  of news source quality.
\newblock {\em Proceedings of the National Academy of Sciences},
  116(7):2521--2526, 2019.

\bibitem{peterson2019human}
Joshua~C Peterson, Ruairidh~M Battleday, Thomas~L Griffiths, and Olga
  Russakovsky.
\newblock Human uncertainty makes classification more robust.
\newblock In {\em Proceedings of the IEEE/CVF International Conference on
  Computer Vision}, pages 9617--9626, 2019.

\bibitem{quoc2013evaluation}
Nguyen Quoc Viet~Hung, Nguyen~Thanh Tam, Lam~Ngoc Tran, and Karl Aberer.
\newblock An evaluation of aggregation techniques in crowdsourcing.
\newblock In {\em International Conference on Web Information Systems
  Engineering}, pages 1--15. Springer, 2013.

\bibitem{raykar2010learning}
Vikas~C Raykar, Shipeng Yu, Linda~H Zhao, Gerardo~Hermosillo Valadez, Charles
  Florin, Luca Bogoni, and Linda Moy.
\newblock Learning from crowds.
\newblock {\em Journal of machine learning research}, 11(4), 2010.

\bibitem{reed2014training}
Scott Reed, Honglak Lee, Dragomir Anguelov, Christian Szegedy, Dumitru Erhan,
  and Andrew Rabinovich.
\newblock Training deep neural networks on noisy labels with bootstrapping.
\newblock {\em arXiv preprint arXiv:1412.6596}, 2014.

\bibitem{rodrigues2017learning}
Filipe Rodrigues, Mariana Lourenco, Bernardete Ribeiro, and Francisco~C
  Pereira.
\newblock Learning supervised topic models for classification and regression
  from crowds.
\newblock {\em IEEE transactions on pattern analysis and machine intelligence},
  39(12):2409--2422, 2017.

\bibitem{rodrigues2018deep}
Filipe Rodrigues and Francisco Pereira.
\newblock Deep learning from crowds.
\newblock In {\em Proceedings of the AAAI Conference on Artificial
  Intelligence}, volume~32, 2018.

\bibitem{rodrigues2014gaussian}
Filipe Rodrigues, Francisco Pereira, and Bernardete Ribeiro.
\newblock Gaussian process classification and active learning with multiple
  annotators.
\newblock In {\em International conference on machine learning}, pages
  433--441. PMLR, 2014.

\bibitem{setio2017validation}
Arnaud Arindra~Adiyoso Setio, Alberto Traverso, Thomas De~Bel, Moira~SN Berens,
  Cas Van Den~Bogaard, Piergiorgio Cerello, Hao Chen, Qi~Dou, Maria~Evelina
  Fantacci, Bram Geurts, et~al.
\newblock Validation, comparison, and combination of algorithms for automatic
  detection of pulmonary nodules in computed tomography images: the luna16
  challenge.
\newblock {\em Medical image analysis}, 42:1--13, 2017.

\bibitem{sheng2017majority}
Victor~S Sheng, Jing Zhang, Bin Gu, and Xindong Wu.
\newblock Majority voting and pairing with multiple noisy labeling.
\newblock {\em IEEE Transactions on Knowledge and Data Engineering},
  31(7):1355--1368, 2017.

\bibitem{smyth1994inferring}
Padhraic Smyth, Usama Fayyad, Michael Burl, Pietro Perona, and Pierre Baldi.
\newblock Inferring ground truth from subjective labelling of venus images.
\newblock {\em Advances in neural information processing systems}, 7, 1994.

\bibitem{song2022learning}
Hwanjun Song, Minseok Kim, Dongmin Park, Yooju Shin, and Jae-Gil Lee.
\newblock Learning from noisy labels with deep neural networks: A survey.
\newblock {\em IEEE Transactions on Neural Networks and Learning Systems},
  2022.

\bibitem{varah1975lower}
James~M Varah.
\newblock A lower bound for the smallest singular value of a matrix.
\newblock {\em Linear Algebra and its applications}, 11(1):3--5, 1975.

\bibitem{wang2021policy}
Jingkang Wang, Hongyi Guo, Zhaowei Zhu, and Yang Liu.
\newblock Policy learning using weak supervision.
\newblock {\em Advances in Neural Information Processing Systems}, 34, 2021.

\bibitem{wang2019symmetric}
Yisen Wang, Xingjun Ma, Zaiyi Chen, Yuan Luo, Jinfeng Yi, and James Bailey.
\newblock Symmetric cross entropy for robust learning with noisy labels.
\newblock In {\em Proceedings of the IEEE/CVF International Conference on
  Computer Vision}, pages 322--330, 2019.

\bibitem{wei2020combating}
Hongxin Wei, Lei Feng, Xiangyu Chen, and Bo~An.
\newblock Combating noisy labels by agreement: A joint training method with
  co-regularization.
\newblock In {\em Proceedings of the IEEE/CVF Conference on Computer Vision and
  Pattern Recognition}, pages 13726--13735, 2020.

\bibitem{wei2021open}
Hongxin Wei, Lue Tao, Renchunzi Xie, and Bo~An.
\newblock Open-set label noise can improve robustness against inherent label
  noise.
\newblock {\em Advances in Neural Information Processing Systems}, 34, 2021.

\bibitem{wei2022deep}
Hongxin Wei, Renchunzi Xie, Lei Feng, Bo~Han, and Bo~An.
\newblock Deep learning from multiple noisy annotators as a union.
\newblock {\em IEEE Transactions on Neural Networks and Learning Systems},
  2022.

\bibitem{wei2021understanding}
Jiaheng Wei, Hangyu Liu, Tongliang Liu, Gang Niu, and Yang Liu.
\newblock Understanding generalized label smoothing when learning with noisy
  labels.
\newblock {\em arXiv preprint arXiv:2106.04149}, 2021.

\bibitem{wei2020optimizing}
Jiaheng Wei and Yang Liu.
\newblock When optimizing $ f $-divergence is robust with label noise.
\newblock {\em arXiv preprint arXiv:2011.03687}, 2020.

\bibitem{wei2021learning}
Jiaheng Wei, Zhaowei Zhu, Hao Cheng, Tongliang Liu, Gang Niu, and Yang Liu.
\newblock Learning with noisy labels revisited: A study using real-world human
  annotations.
\newblock {\em arXiv preprint arXiv:2110.12088}, 2021.

\bibitem{whitehill2009whose}
Jacob Whitehill, Ting-fan Wu, Jacob Bergsma, Javier Movellan, and Paul Ruvolo.
\newblock Whose vote should count more: Optimal integration of labels from
  labelers of unknown expertise.
\newblock {\em Advances in neural information processing systems}, 22, 2009.

\bibitem{xia2020robust}
Xiaobo Xia, Tongliang Liu, Bo~Han, Chen Gong, Nannan Wang, Zongyuan Ge, and
  Yi~Chang.
\newblock Robust early-learning: Hindering the memorization of noisy labels.
\newblock In {\em International conference on learning representations}, 2020.

\bibitem{xia2020part}
Xiaobo Xia, Tongliang Liu, Bo~Han, Nannan Wang, Mingming Gong, Haifeng Liu,
  Gang Niu, Dacheng Tao, and Masashi Sugiyama.
\newblock Part-dependent label noise: Towards instance-dependent label noise.
\newblock {\em Advances in Neural Information Processing Systems},
  33:7597--7610, 2020.

\bibitem{xia2019anchor}
Xiaobo Xia, Tongliang Liu, Nannan Wang, Bo~Han, Chen Gong, Gang Niu, and
  Masashi Sugiyama.
\newblock Are anchor points really indispensable in label-noise learning?
\newblock {\em Advances in Neural Information Processing Systems}, 32, 2019.

\bibitem{yu2019does}
Xingrui Yu, Bo~Han, Jiangchao Yao, Gang Niu, Ivor Tsang, and Masashi Sugiyama.
\newblock How does disagreement help generalization against label corruption?
\newblock In {\em International Conference on Machine Learning}, pages
  7164--7173. PMLR, 2019.

\bibitem{zhou2012ensemble}
Zhi-Hua Zhou.
\newblock {\em Ensemble methods: foundations and algorithms}.
\newblock CRC press, 2012.

\bibitem{zhu2022detect}
Zhaowei Zhu, Zihao Dong, and Yang Liu.
\newblock Detecting corrupted labels without training a model to predict.
\newblock {\em arXiv preprint arXiv:2110.06283}, 2022.

\bibitem{zhu2021second}
Zhaowei Zhu, Tongliang Liu, and Yang Liu.
\newblock A second-order approach to learning with instance-dependent label
  noise.
\newblock In {\em Proceedings of the IEEE/CVF Conference on Computer Vision and
  Pattern Recognition}, pages 10113--10123, 2021.

\bibitem{zhu2021rich}
Zhaowei Zhu, Tianyi Luo, and Yang Liu.
\newblock The rich get richer: Disparate impact of semi-supervised learning.
\newblock {\em arXiv preprint arXiv:2110.06282}, 2021.

\bibitem{zhu2021clusterability}
Zhaowei Zhu, Yiwen Song, and Yang Liu.
\newblock Clusterability as an alternative to anchor points when learning with
  noisy labels.
\newblock In {\em International Conference on Machine Learning}, pages
  12912--12923. PMLR, 2021.

\bibitem{zhu2022beyond}
Zhaowei Zhu, Jialu Wang, and Yang Liu.
\newblock Beyond images: Label noise transition matrix estimation for tasks
  with lower-quality features.
\newblock {\em arXiv preprint arXiv:2202.01273}, 2022.

\end{thebibliography}
